\documentclass{article} 
\usepackage{iclr2027_conference,times}
\iclrfinalcopy

\usepackage{amsmath,amsfonts,bm}

\def\eqref#1{equation~\ref{#1}}

\def\1{\bm{1}}

\DeclareMathAlphabet{\mathsfit}{\encodingdefault}{\sfdefault}{m}{sl}
\SetMathAlphabet{\mathsfit}{bold}{\encodingdefault}{\sfdefault}{bx}{n}

\usepackage{hyperref}
\usepackage{url}
\usepackage{graphicx}
\usepackage{booktabs}
\usepackage{amsmath}
\usepackage{amssymb}
\usepackage{multirow}
\usepackage{array}
\usepackage{algorithm}
\usepackage{algpseudocode}
\usepackage{xcolor}
\usepackage{colortbl}
\usepackage{listings}
\usepackage{float}
\usepackage{placeins}

\definecolor{chemopdblue}{HTML}{B8BBE3}
\definecolor{tableink}{HTML}{27324A}
\definecolor{casepromptbg}{HTML}{FAFAFB}
\definecolor{casepromptedge}{HTML}{B9BDC7}
\definecolor{casechemopdbg}{HTML}{F1F2FA}
\definecolor{casechemopdedge}{HTML}{6872A8}
\definecolor{casebaselinebg}{HTML}{FBF4F5}
\definecolor{casebaselineedge}{HTML}{B18491}
\newcolumntype{L}[1]{>{\raggedright\arraybackslash}m{#1}}
\newcolumntype{R}[1]{>{\raggedleft\arraybackslash}m{#1}}
\newcolumntype{C}[1]{>{\centering\arraybackslash}m{#1}}
\newlength{\panelmethodwidth}
\newlength{\panelmetricwidth}
\newlength{\panelconfigwidth}
\newlength{\panelmodelwidth}
\newcommand{\twolineheight}{\vphantom{\shortstack[c]{X\\X}}}

\newcommand{\method}{ChemOPD}
\newcommand{\taskset}{\mathcal{T}}
\newcommand{\groupset}{\mathcal{G}}

\lstdefinestyle{casetranscript}{
  basicstyle=\ttfamily\footnotesize,
  breaklines=true,
  breakatwhitespace=false,
  columns=fullflexible,
  keepspaces=true,
  showstringspaces=false,
  frame=single,
  framerule=0.45pt,
  framesep=4pt,
  xleftmargin=1pt,
  xrightmargin=1pt,
  aboveskip=3pt,
  belowskip=7pt
}
\lstdefinestyle{caseprompt}{
  style=casetranscript,
  backgroundcolor=\color{casepromptbg},
  rulecolor=\color{casepromptedge}
}
\lstdefinestyle{casechemopd}{
  style=casetranscript,
  backgroundcolor=\color{casechemopdbg},
  rulecolor=\color{casechemopdedge}
}
\lstdefinestyle{casebaseline}{
  style=casetranscript,
  backgroundcolor=\color{casebaselinebg},
  rulecolor=\color{casebaselineedge}
}

\title{ChemOPD: Multi-Teacher On-Policy Distillation for Multi-Task Chemical Reasoning}

\author{Yaoyao Xu\textsuperscript{$\S$ \P }, 
Xinjian Zhao\textsuperscript{$\S$ \P }, 
Xiaozhuang Song\textsuperscript{$\S$ \P}, 
Xuemin Chen\textsuperscript{$\S$ \P},
Tianshu Yu\textsuperscript{$\S$ \P }
\\
\textsuperscript{$\S$} School of Data Science, The Chinese University of Hong Kong, Shenzhen \\
\textsuperscript{\P} Shanghai Artificial Intelligence Laboratory\\
 \texttt{\{yaoyaoxu,xinjianzhao1,xiaozhuangsong1,xueminchen\}@link.cuhk.edu.cn}, \\
 \texttt{yutianshu@cuhk.edu.cn} 
}

\iclrfinalcopy 
\begin{document}

\maketitle

\begin{abstract}
Large language models are increasingly expected to support diverse chemical
reasoning capabilities within a unified model. One approach is to develop
specialized capabilities separately and consolidate them through multi-teacher
on-policy distillation, but this raises two questions: how should
specialization be organized, and how should specialist guidance be integrated?
We introduce \method{}, which addresses both. We estimate task affinities from
supervised fine-tuning gradients and solve a constrained mixed-integer program
(MIP) to construct partially overlapping specialist groups. During distillation, we retain a generalist teacher trained on all tasks
so that specialist guidance supplements rather than replaces its supervision.
Our anchor-residual objective gradually increases the routed specialist's
contribution on student-generated responses. On ChemCoTBench, affinity-guided
specialization produces task-dependent gains over the generalist teacher
and improves several capabilities beyond semantic task grouping. Yet stronger teacher-side
performance does not automatically yield stronger students: with the same
specialists and routes, anchor-residual OPD improves most reported metrics
over specialist-only distillation and realizes a larger share of the available
teacher gains.  These results highlight specialization and capability
integration as connected but distinct design problems in chemical reasoning. 
\end{abstract}

\section{Introduction}

Large language models (LLMs) are increasingly being developed to solve a
broad range of chemical problems. Beyond chemical question answering, recent
models have been applied to molecular property prediction, molecule generation
and editing, reaction prediction, retrosynthesis, and multistep chemical
reasoning
\citep{yu2024llasmol,zhao2025chemdfm,lin2025enhancing,zhao2026molemb,liu2026retro,hao2026beyond,song2026aot,wang2026survey}.
These tasks rely on overlapping chemical knowledge, but demand different
forms of reasoning and different outputs: predicting a property, modifying a
molecular structure, or inferring a reaction outcome does not require exactly
the same capability. As chemical language models become more general, an
important challenge is therefore not only to improve individual tasks, but to
develop a broad collection of chemical capabilities within a single model.

A single model at inference does not require every capability to be developed
through the same training process. Knowledge distillation provides a natural
way to separate capability development from capability deployment: stronger
or more specialized models can acquire useful behaviors during training, and
their knowledge can later be transferred into a unified student
\citep{hinton2015distilling,xu2024survey}. This idea is particularly appealing when
different tasks benefit from different forms of specialization. Multi-task
learning has long shown that sharing data and parameters across related tasks
can be beneficial, while task-grouping studies further demonstrate that which
tasks are learned together can strongly affect the capabilities that emerge
\citep{standley2020tasks,fifty2021efficiently, song2022efficient}.
Thus, developing specialists is not simply a matter of assigning one model to
each task: it also requires deciding which tasks should contribute jointly to
each specialist.

Once specialized capabilities have been developed, they must still be made
useful to a shared student. For autoregressive models, on-policy distillation
(OPD) lets a teacher supervise the contexts that the student actually visits
during generation, rather than only imitating completed teacher solutions
\citep{agarwal2024policy,gu2024minillm,song2026survey,li2026rethinking,xu2026tip}. Multi-teacher OPD further enables
capabilities developed in different teachers to be consolidated into one
student \citep{ma2026mopd}. This provides a natural route from specialization
to a unified model, but leaves two choices unresolved for heterogeneous
chemical reasoning:
\emph{how should specialization itself be organized, and how should the
resulting specialist guidance be combined when training a shared student?}

The first question concerns how chemical tasks should be grouped when
developing specialists. Human-readable task categories describe semantic
differences, but they need not fully reflect how training one task affects the
learning of another. We therefore ground specialist construction in measured
training relationships. Using supervised fine-tuning (SFT) gradients, we estimate local task
affinities \citep{fifty2021efficiently}. Building on optimization-based task grouping \citep{kang2011learning, wang2024towards}, we adapt a constrained mixed-integer formulation to construct a small collection
of partially overlapping specialists.
The overlap is intentional: a task can contribute useful training signal to
a specialist even when that specialist is not ultimately responsible for
supervising the task during distillation. In this way, task relationships
inform how expertise is developed without requiring benchmark categories to
serve as fixed training boundaries.

The second question concerns how specialist expertise should enter the
student's learning process. Rather than rely only on routed specialists,
we retain a generalist teacher trained on all tasks so that task-specific
guidance supplements its supervision. Our anchor-residual OPD objective
uses the generalist's guidance as the anchor and adds a correction from
the routed specialist on the same student-generated prefixes. We gradually
increase the specialist weight through a linear warm-up. At inference,
only the unified student is used.

We evaluate \method{} on ChemCoTBench
\citep{hao2026beyond}, separating the quality of the constructed teachers from
the student's ability to benefit from them. Affinity-guided grouping produces
task-dependent specialist advantages over the generalist teacher and improves
several capabilities relative to benchmark-family grouping. However, stronger
teacher-side performance does not automatically translate into stronger
students: positive specialist headroom can coexist with poor or even negative
student gains under specialist-only OPD. With the same specialist checkpoints
and task routes, anchor-residual OPD realizes a broader fraction of the
available teacher gains and improves most reported metrics over
specialist-only distillation. Component ablations further support combining generalist supervision with gradual specialist weighting, rather than using specialists alone or applying the final mixture throughout training. Together, these
results suggest that developing specialized capabilities and making them
usable by a unified student are connected but distinct design problems.

Our contributions are:
\begin{itemize}
    \item We develop affinity-guided specialist construction for heterogeneous
    chemical reasoning, using measured task interactions and constrained
    grouping to organize partially shared specialist training.

    \item We introduce anchor-residual on-policy distillation, which treats
    routed specialist guidance as a correction to a generalist teacher's
    supervision and gradually increases its contribution during training.

    \item We separate teacher-side capability from realized student gains and
    show that effective capability integration depends on both the source and
    the training-time balance of shared and specialist supervision.
\end{itemize}

\begin{figure*}[ht]
\centering
\IfFileExists{figure/OPD_framework.pdf}{%
  \includegraphics[width=\textwidth]{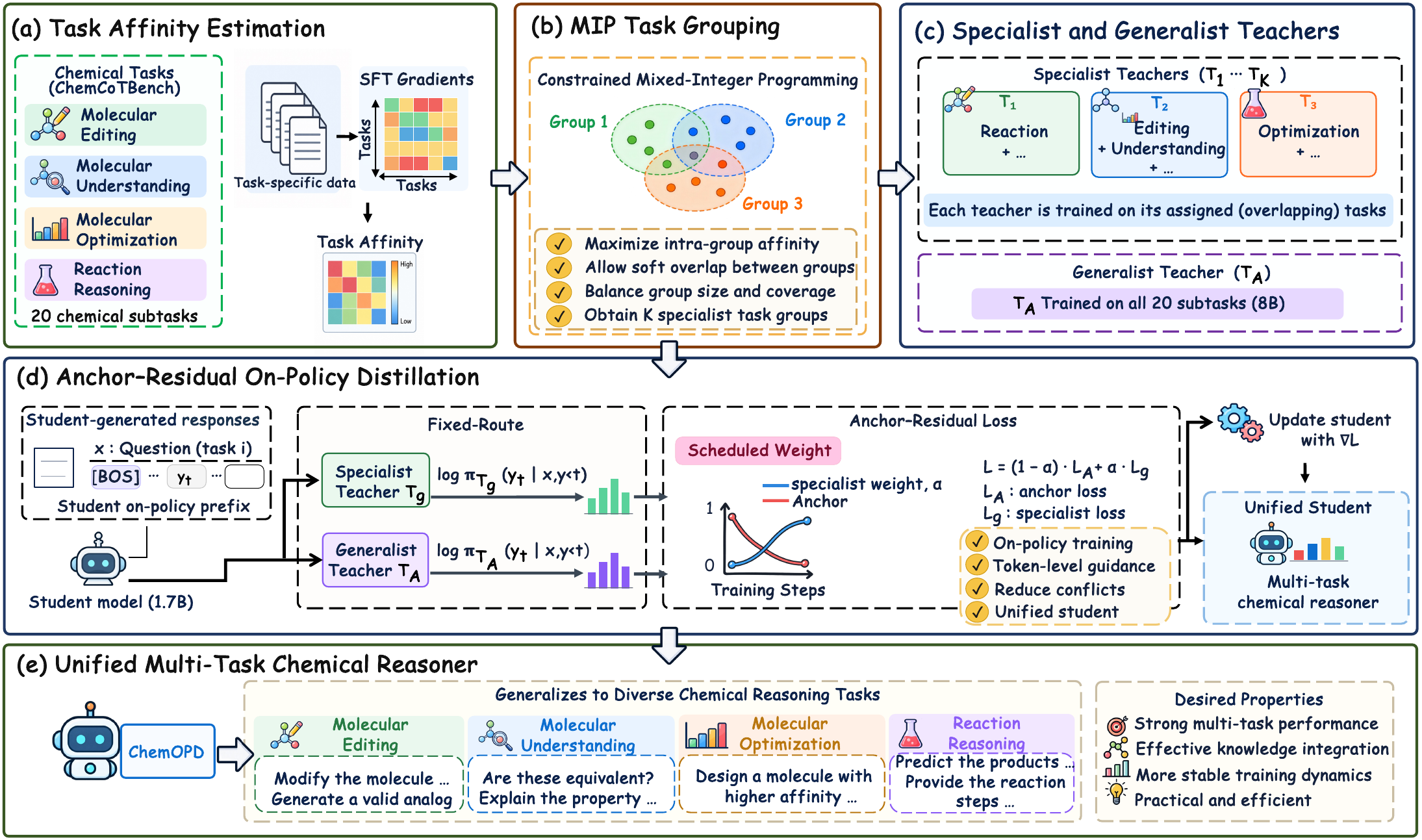}%
}{%
  \includegraphics[width=\textwidth]{figure/method_framework_v3_candidate3_prefix_workbench.png}%
}\vspace{-1mm}
\caption{Overview of affinity-guided teacher construction and anchor-residual
multi-teacher OPD. The MIP converts gradient-based task affinities into
overlapping task groups for specialist training. During OPD, the generalist
teacher and the fixed-route specialist score the same student-generated
prefixes. The generalist's advantage serves as the anchor, and the specialist
residual is introduced gradually to train a unified student.}
\label{fig:overview}
\end{figure*}

\section{Related Work}
\noindent\textbf{Chemical reasoning with language models.}
Large language models have been adapted across chemical prediction,
molecular generation, and reasoning. Fine-tuned LLMs have been studied for
molecular and materials properties, reaction yields, and inverse design
\citep{jablonka2024leveraging}, while MolT5 connects molecular structures
with natural language through molecule captioning and text-conditioned
generation \citep{edwards2022translation}. Large-scale instruction-tuning
efforts such as Mol-Instructions and LlaSMol further cover diverse molecular,
biomolecular, and reaction-related tasks
\citep{fang2024mol,yu2024llasmol}. Chemistry-specialized models
including ChemLLM and ChemDFM pursue broader domain capabilities within a
single language model \citep{zhang2024chemllm,zhao2025developing}, while
ether0 explicitly studies reasoning-oriented post-training across diverse
chemistry problems \citep{narayanan2026training}. ChemCoTBench complements
these model-development efforts with structured evaluation of molecular
understanding, editing, optimization, and reaction reasoning
\citep{hao2026beyond}. Our work focuses on how such heterogeneous chemical
capabilities can be developed through specialization and subsequently
integrated into a unified student.

\noindent\textbf{Task grouping in multi-task learning.}
Multi-task learning has long studied which tasks should share training.
Prior work considers partially overlapping task groups
\citep{kumar2012learning}, searches for effective task combinations
\citep{standley2020tasks}, and estimates task affinity from cross-task
training effects \citep{fifty2021efficiently}. Constrained optimization has
also been used to form task groups under coverage and resource constraints
\citep{wang2024towards}. We build on this line of work to organize the
training mixtures of specialist teachers before distillation.

\noindent\textbf{On-policy and multi-teacher distillation.}
Knowledge distillation transfers teacher behavior to a student
\citep{hinton2015distilling}. For language models, MiniLLM and GKD study
distillation with reverse-KL and student-generated sequences
\citep{gu2024minillm,agarwal2024policy}. MOPD extends on-policy distillation
to multiple domain-specific teachers that are consolidated into a unified
student \citep{ma2026mopd}, with recent work exploring other forms of
multi-teacher supervision
\citep{wang2026mad,chen2026counteraction,zhou2026danceopd}.

\section{Problem Setting}
Let $\taskset=\{1,\ldots,n\}$ denote a collection of chemical subtasks and
let $D_i$ be the SFT data for task $i$. We seek $m$ teacher groups
$\groupset=\{G_1,\ldots,G_m\}$. Every task must appear in at least one
group; group-size and membership constraints limit overlap. Training a
teacher on $\cup_{i\in G_g}D_i$ yields specialist $T_g$, while a generalist
teacher $T_A$ is trained on $\cup_i D_i$.
The student $\pi_{\theta_0}$ is initialized by all-task SFT. Each OPD prompt
$x$ has a known subtask label $c(x)$, and a routing function
$r(c(x))\in\{1,\ldots,m\}$ selects one specialist. Although teacher training
groups may overlap, each routed example receives one specialist route,
fixed before OPD. The generalist-only baseline has no specialist
route. We evaluate both whether the groups produce useful specialist
advantages and whether these advantages can be incorporated into a student
with broad cross-task capability.

\section{Method}

\method{} addresses teacher construction and capability integration in two
stages. Stage I uses SFT-gradient affinities and a constrained MIP to select
partially overlapping training groups for specialist teachers. Stage II
combines the generalist teacher's supervision with a residual correction
from the routed specialist on student-generated responses. A linear warm-up
gradually increases the specialist contribution.
Figure~\ref{fig:overview} summarizes the two stages.

\subsection{Gradient Probes and Task Affinity}

To construct specialists, we estimate whether training on one task could
locally benefit another \citep{fifty2021efficiently}. Each subtask supplies
two disjoint probe sets drawn from its SFT training data: a source-update
probe and a target-loss probe, referred to as the training and validation
probes below.

Let $\phi$ denote the model parameters and $\phi_0$ the teacher's initial
checkpoint. For task $i$, let $L_i^{\mathrm{tr}}(\phi)$ and
$L_i^{\mathrm{val}}(\phi)$ denote the SFT losses on its training and
validation probes, respectively. Each loss averages cross-entropy over
valid assistant tokens within each example, then equally across examples.
For a source task $i$ and a target task $j$, let
$\bar g_i^{\mathrm{tr}}$ and $\bar g_j^{\mathrm{val}}$ denote the
training-probe gradient of $L_i^{\mathrm{tr}}$ and the validation-probe
gradient of $L_j^{\mathrm{val}}$. Both are evaluated at $\phi_0$, with
coordinates outside a common probed parameter subset set to zero.
For a small step size $\eta>0$, a first-order expansion gives
\begin{equation}
L_j^{\mathrm{val}}(\phi_0-\eta\bar g_i^{\mathrm{tr}})
\approx
L_j^{\mathrm{val}}(\phi_0)
-\eta\left\langle
\bar g_i^{\mathrm{tr}},\bar g_j^{\mathrm{val}}
\right\rangle.
\label{eq:affinity-first-order}
\end{equation}
A positive inner product therefore predicts a local reduction in task
$j$'s loss. To make comparisons tractable, we map the retained gradients
to a common $d_{\mathrm{CS}}$-dimensional space using a fixed CountSketch
projection \citep{charikar2002finding}. The projection is shared across
tasks and probe splits; write
$\tilde g=\operatorname{CS}(\bar g)\in\mathbb{R}^{d_{\mathrm{CS}}}$.

For each ordered task pair $(i,j)$, let $D_{ij}$ denote the gradient
alignment and $A_{ij}$ its target-loss-normalized form. These serve as
local proxies for absolute and relative loss reduction per unit step
size, respectively. Let $C_{ij}$ denote the cosine affinity, which measures agreement
between gradient directions~\citep{du2018adapting}. With a small constant $\epsilon>0$ for numerical stability, we compute
these quantities from the compressed gradients:
\begin{equation}
D_{ij}=
\langle\tilde g_i^{\mathrm{tr}},\tilde g_j^{\mathrm{val}}\rangle,
\quad
A_{ij}=
\frac{D_{ij}}{\max(L_j^{\mathrm{val}}(\phi_0),\epsilon)},
\quad
C_{ij}=
\frac{D_{ij}}
{\max(\|\tilde g_i^{\mathrm{tr}}\|
      \|\tilde g_j^{\mathrm{val}}\|,\epsilon)}.
\label{eq:task-affinity}
\end{equation}
We use the cosine affinity matrix $C=[C_{ij}]$ for grouping to reduce
the influence of gradient-scale differences across tasks. These affinities
are local training proxies; the resulting groups are assessed through
teacher and student performance. Probe sizes, parameter selection, and the
sketch dimension are specified in Appendix~\ref{app:groups}.

\subsection{MIP Grouping and Specialist Teachers}

We use a constrained MIP to turn task affinities into $m$ partially
overlapping training groups, balancing within-group compatibility, coverage,
and sharing~\citep{wang2024towards}.

Let $S$ denote the standardized task-affinity matrix, and let $\mu$ and
  $\sigma$ denote the population mean and standard deviation of all
  $n(n-1)$ directed off-diagonal entries of $C$. We compute
  $S_{ij}=(C_{ij}-\mu)/\sigma$ for $i\ne j$ and set $S_{ii}=0$.
  We then define the symmetric task compatibility as $W_{ij}=S_{ij}+S_{ji}$, which combines the two directed affinities.

Let $x_{ig}\in\{0,1\}$ indicate whether task $i$ belongs to group $g$,
and let $k_g$ denote the number of tasks in that group.
Let $q_{\max}$ be the maximum number of groups per task and
$\lambda\ge0$ the weight of the membership penalty.
Let $k_{\min}$ and $k_{\max}$ be the lower and upper group-size bounds.
We enumerate feasible size profiles $(k_1,\ldots,k_m)$ satisfying these
bounds. For each profile, we solve
\begin{equation}
\begin{aligned}
\max_x \quad &
\sum_{g=1}^{m}\frac{1}{k_g}
\sum_{i<j}W_{ij}x_{ig}x_{jg}
-\lambda\sum_{i,g}x_{ig}\\
\text{s.t.}\quad &
\sum_i x_{ig}=k_g \quad(\forall g), \qquad
1\leq\sum_g x_{ig}\leq q_{\max} \quad(\forall i).
\end{aligned}
\label{eq:mip}
\end{equation}
The first term rewards within-group compatibility normalized by group
size. The second penalizes total task--group memberships.
This penalty is constant within a fixed profile and discourages
additional memberships when comparing profiles.
The constraints enforce group sizes and task coverage with limited
overlap. We linearize the binary products using auxiliary variables
and solve each profile with SciPy/HiGHS~\citep{huangfu2018parallelizing}.
We retain the returned feasible assignment with the highest objective
across profiles.

Each selected group defines the SFT data for specialist $T_g$. Group
membership determines which tasks it learns from, whereas the fixed route
determines which tasks it supervises during OPD. Overlap thus permits sharing
during teacher training while retaining one specialist route per prompt.
The generalist teacher $T_A$ uses the same initialization and SFT recipe
on all tasks. Its supervision is used alone in Vanilla OPD and as the
anchor in ChemOPD. Appendix~\ref{app:groups} provides the algorithm,
configuration, memberships, and routes.

\subsection{Routed and Anchor-Residual OPD}

Let $\pi_\theta$ denote the student policy, and let $\pi_{\phi_g}$
and $\pi_{\phi_A}$ denote the frozen policies of specialist $T_g$
and generalist teacher $T_A$, respectively.
We draw a prompt $x$ and its subtask label $c$ from the OPD data
distribution $\mathcal D_{\mathrm{OPD}}$, generate a student response
$y\sim\pi_\theta(\cdot\mid x)$, and select specialist $g=r(c)$
using the fixed route.
Let $\mathcal M(x,y)$ denote the valid response positions and
$h_t$ the prefix $(x,y_{<t})$ at position $t$.
The routed objective $\mathcal L_{\mathrm{rev\text{-}KL}}^{(g)}$
measures the expected per-token reverse KL between the student
and its specialist~\citep{ma2026mopd}:
\begin{equation}
\mathcal L_{\mathrm{rev\text{-}KL}}^{(g)}(\theta)
=
\mathbb E_{(x,c),\,y\sim\pi_\theta}
\left[
\frac{1}{|\mathcal M(x,y)|}
\sum_{t\in\mathcal M(x,y)}
D_{\mathrm{KL}}\!\left(
\pi_\theta(\cdot\mid h_t)\,\|\,
\pi_{\phi_g}(\cdot\mid h_t)
\right)
\right].
\label{eq:routed-opd}
\end{equation}

For the implemented actor update, we use a top-$K$ teacher-derived
advantage \citep{ma2026mopd,gu2024minillm}, where $K$ is the number of
student-selected token candidates scored at each prefix.
Let $v_{t,k}$ denote the student's $k$-th most probable token at
prefix $h_t$, and let $\bar\pi_{\theta,t,k}$ denote its probability
normalized over these candidates.
Let $\hat A_{g,t,k}$ denote the specialist-derived advantage
that weights the actor update for this candidate.
With $\operatorname{sg}$ denoting the stop-gradient operator,
we compute
\begin{equation}
\bar\pi_{\theta,t,k}
=
\frac{\pi_\theta(v_{t,k}\mid h_t)}
{\sum_{j=1}^{K}\pi_\theta(v_{t,j}\mid h_t)},
\qquad
\hat A_{g,t,k}
=
\operatorname{sg}\!\left[
\bar\pi_{\theta,t,k}
\log\frac{\pi_{\phi_g}(v_{t,k}\mid h_t)}
{\pi_\theta(v_{t,k}\mid h_t)}
\right].
\label{eq:topk-teacher-signal}
\end{equation}
All model probabilities use temperature-scaled full-vocabulary
softmax distributions; only the student weights
$\bar\pi_{\theta,t,k}$ are normalized over the selected candidates.

To combine generalist and specialist supervision, we score the same
prefixes and candidates with both teachers. The generalist's advantage
$\hat A_{A,t,k}$ serves as the anchor and is computed by replacing
$\pi_{\phi_g}$ with $\pi_{\phi_A}$ in
Equation~\ref{eq:topk-teacher-signal}.
At OPD update $s$, let $\alpha_g(s)\in[0,1]$ denote the specialist
weight, shared across response positions and candidates.
The combined advantage $\hat A_{\mathrm{ChemOPD},t,k}(s)$
adds a weighted specialist correction to the anchor:
\begin{equation}
\hat A_{\mathrm{ChemOPD},t,k}(s)
=
\hat A_{A,t,k}
+\alpha_g(s)\bigl(\hat A_{g,t,k}-\hat A_{A,t,k}\bigr).
\label{eq:anchor-residual}
\end{equation}
The specialist residual $\hat A_{g,t,k}-\hat A_{A,t,k}$ adjusts the
generalist's guidance where the two teachers differ in candidate probabilities.
Setting $\alpha_g=0$ or $1$ recovers generalist-only or specialist-only
supervision, respectively.
We use the combined advantage in the on-policy actor surrogate specified
in Appendix~\ref{app:opd-training-diagnostics}. In the reported
configuration, this reduces locally to an advantage-weighted score-function
update, with the clipping terms numerically inactive.

We linearly increase the specialist weight to introduce this correction
gradually. Let $U\geq2$ denote the total number of OPD updates and
$\alpha_g^{\max}$ the maximum specialist weight.
For a warm-up fraction $\rho\in(0,1]$, let $U_{\mathrm{warm}}$
denote the warm-up duration in updates. For $s=1,\ldots,U$, we use
\begin{equation}
U_{\mathrm{warm}}=\max\{2,\lceil\rho U\rceil\},
\qquad
\alpha_g(s)=
\alpha_g^{\max}
\min\!\left\{1,\frac{s-1}{U_{\mathrm{warm}}-1}\right\}.
\label{eq:alpha-schedule}
\end{equation}
Without warm-up ($\rho=0$), we set
$\alpha_g(s)=\alpha_g^{\max}$ throughout training.
The specialist weight controls both the balance between teacher signals
and their exposure over training. The reported values and candidate support
size are specified in Appendices~\ref{app:chemopd-ablation}
and~\ref{app:implementation}.

\section{Experiments}
\label{sec:experiments}
\label{sec:results}
\label{sec:analysis}

\subsection{Experimental Setup}
\label{sec:experimental-setup}

We evaluate \method{} on 20 ChemCoTBench subtasks spanning molecular editing,
understanding, optimization, and reaction reasoning \citep{hao2026beyond}.
We use the released structured-CoT training data and task-specific metrics.
Only the retrosynthesis evaluation split is changed: malformed references
motivate a product-string-disjoint holdout from the training pool. This is
not a canonical chemical-identity split; Appendix~\ref{app:tasks} gives the
data and scoring rules.
We train three Qwen3-8B specialists and an additional Qwen3-8B generalist
teacher, and distill them into a Qwen3-1.7B student
\citep{yang2025qwen3}. Teachers share their initialization and SFT recipe;
the student starts from all-task SFT. ChemOPD routes each prompt to one specialist and combines its guidance with the generalist anchor. Appendices ~\ref{app:implementation} and ~\ref{app:chemopd-ablation} provide the implementation details and ablations.
We compare MIP specialists with the generalist and with specialists
trained on the benchmark's semantic task families. For student integration,
Vanilla OPD uses only the generalist; Benchmark-family OPD uses benchmark-family
specialists; and MOPD shares ChemOPD's specialists, routes, student
initialization, prompt stream, and actor-update budget but uses specialists
alone \citep{ma2026mopd}.
Random-group OPD keeps the anchor-residual rule and changes the specialist
collection. Homogeneous-RR and the DanceOPD-style control vary teacher
sampling and batch construction. Full comparison protocols are in
Appendix~\ref{app:comparison-protocols}.

\subsection{Constructing Specialist Teachers}
\label{sec:analysis-grouping}

\begin{figure*}[ht]
\centering
\includegraphics[width=\textwidth]{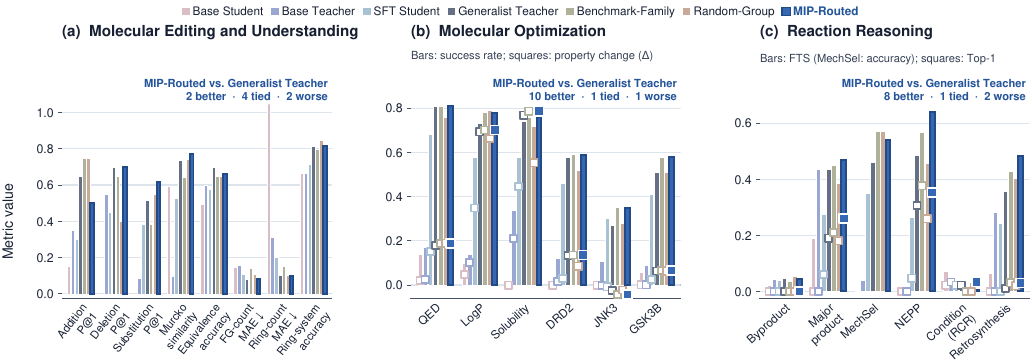}
 \caption{Performance of teacher configurations across benchmark tasks.
  Base Student and Base Teacher are the pretrained Qwen3-1.7B and Qwen3-8B
  models, respectively, while SFT Student is the common all-task SFT
  initializer. Generalist Teacher is trained on all tasks, whereas
  Benchmark-Family, Random-Group, and MIP-Routed denote three specialist-routing
  schemes. Bars show the primary metric for each task. Squares show property
  change ($\Delta$) in panel (b) and Top-1 in panel (c); MechSel is reported by
  accuracy. Downward arrows mark lower-is-better metrics. Blue annotations
  summarize the better/tied/worse counts of MIP-Routed relative to Generalist
  Teacher. Full results are provided in Appendix
  Table~\ref{tab:teacher-results-full}.}
\label{fig:teacher-grouping-outcomes}
\end{figure*}

Affinity-guided grouping produces task-dependent specialist advantages.
The MIP-routed teacher improves 20 of the 31 reported metrics over the
generalist teacher, including retrosynthesis FTS from 0.358 to 0.482
(Figure~\ref{fig:teacher-grouping-outcomes}). Relative to semantic grouping,
optimization scores are unchanged, while several reaction scores improve.
Human-defined groupings likewise show gains on some tasks and losses on
others (Appendix~\ref{app:teacher-results}). These comparisons support using
training relationships to refine specialist task mixtures, rather than
expecting specialization to improve every task.

\begin{figure*}[ht]
\centering
\includegraphics[width=0.98\textwidth]{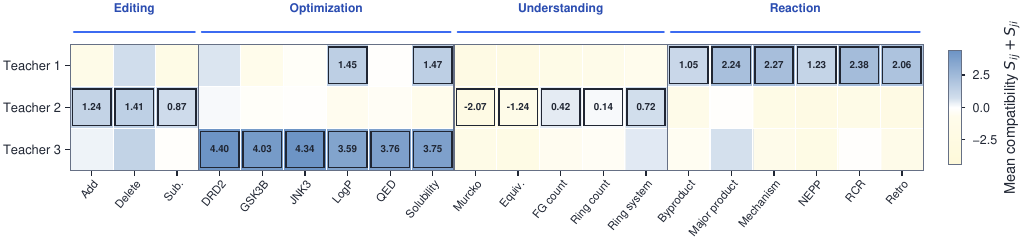}\vspace{-3.9mm}
 \caption{Affinity-based task--group compatibility for the selected MIP
  grouping. Each cell reports the mean compatibility $W_{ij}$ between the
  column task and the other tasks assigned to the row teacher. Outlined cells
  indicate the memberships selected by the MIP. LogP and solubility are shared
  by Teachers~1 and~3.}
\label{fig:mip-compatibility}
\end{figure*}

\noindent \textbf{Task organization.}
The groups contain 8, 8, and 6 tasks (Figure~\ref{fig:mip-compatibility}).
Editing and understanding share a specialist. LogP and solubility appear in
both reaction and optimization training groups but are routed only to the
optimization specialist during OPD. The reaction teacher therefore learns
from tasks beyond those it supervises, illustrating the distinction between
training membership and teaching assignments. Coverage and capacity also
constrain the grouping; pairwise compatibility is not imposed as a hard
requirement. Appendix~\ref{app:groups} provides the full memberships and
pairwise affinities.
Ten probe draws recover the same grouping up to teacher-label
permutation, including both overlap tasks. This supports repeatability of
the grouping decision under probe resampling, rather than downstream
training variability (Appendix~\ref{app:affinity-robustness}).

\subsection{Integrating Specialist Capabilities}
\label{sec:analysis-opd}

\noindent \textbf{Student performance.}
ChemOPD improves most reported metrics over Vanilla OPD and Benchmark-family
OPD, with gains in editing, optimization, and reaction reasoning
(Table~\ref{tab:detailed-opd-results}). With specialists and routes held
fixed, it also matches or improves on MOPD in 26 of 31 metrics. Solubility
improvement reaches 0.718, compared with 0.490 under generalist-only Vanilla OPD
and 0.506 under specialist-only MOPD. The combined training rule can thus
outperform either source used alone, although individual tasks still favor
single-source supervision.
Random-group OPD retains the same integration rule but performs worse on
most metrics, while remaining stronger on LogP. This contrast supports
considering teacher construction alongside the integration rule.
Homogeneous-RR and DanceOPD-style sampling are competitive on some editing
and optimization metrics but generally weaker on reaction reasoning.
Changing how teachers are sampled therefore produces different task
trade-offs, rather than reproducing the same gains.

\begin{table*}[ht]
\caption{Task-level results across OPD methods. Each panel
retains the native metric for each subtask; bold marks the best OPD method in a column before display rounding, and the light-blue row highlights \method. FG and Ring are lower-is-better; all other metrics are higher-is-better.}
\label{tab:detailed-opd-results}
\centering
\begingroup
\footnotesize
\renewcommand{\arraystretch}{1.08}
\setlength{\tabcolsep}{0pt}

\makebox[\textwidth][c]{{\small\textcolor{tableink}{\bfseries (a) Molecule editing \& understanding}}}\par\vspace{1pt}
\setlength{\panelmethodwidth}{3.50cm}
\setlength{\panelmetricwidth}{\textwidth}
\addtolength{\panelmetricwidth}{-\panelmethodwidth}
\divide\panelmetricwidth by 8
\begin{tabular}{@{}L{\panelmethodwidth}*{8}{C{\panelmetricwidth}}@{}}
\toprule
& \multicolumn{3}{c}{\textsc{Editing}} & \multicolumn{5}{c}{\textsc{Understanding}} \\
\cmidrule(lr){2-4}\cmidrule(lr){5-9}
\textbf{Method} &
\shortstack[c]{Add\\P@1} &
\shortstack[c]{Delete\\P@1} &
\shortstack[c]{Sub.\\P@1} &
\shortstack[c]{Murcko\\Sim.} &
\shortstack[c]{Equiv.\\Acc.} &
\shortstack[c]{FG\\MAE $\downarrow$} &
\shortstack[c]{Ring\\MAE $\downarrow$} &
\shortstack[c]{Ring-Sys\\Sim.} \\
\midrule
Student, all-task SFT & 0.30 & 0.45 & 0.38 & 0.53 & 0.58 & 0.11 & 0.20 & 0.72 \\
MIP-routed SFT  & 0.50 & 0.70 & 0.62 & 0.77 & 0.66 & 0.08 & 0.10 & 0.82 \\
\midrule
Vanilla OPD & 0.35 & 0.55 & 0.48 & 0.55 & 0.55 & \textbf{0.07} & \textbf{0.15} & 0.77 \\
Benchmark-family OPD & 0.45 & 0.25 & 0.40 & 0.53 & 0.60 & 0.10 & \textbf{0.15} & 0.70 \\
MOPD & \textbf{0.50} & 0.20 & 0.52 & 0.48 & 0.61 & 0.14 & 0.25 & 0.77 \\
Homogeneous-RR & 0.20 & 0.55 & 0.08 & 0.47 & \textbf{0.67} & 0.12 & 0.20 & 0.78 \\
DanceOPD & 0.35 & 0.65 & 0.60 & 0.52 & 0.64 & 0.11 & \textbf{0.15} & 0.73 \\
Random-group OPD & 0.30 & 0.60 & 0.65 & 0.53 & 0.56 & 0.09 & \textbf{0.15} & 0.75 \\
\rowcolor{chemopdblue!20}\textbf{ChemOPD} & 0.35 & \textbf{0.75} & \textbf{0.70} & \textbf{0.58} & 0.62 & 0.10 & \textbf{0.15} & \textbf{0.83} \\
\bottomrule
\end{tabular}

\par\vspace{4pt}
\makebox[\textwidth][c]{{\small\textcolor{tableink}{\bfseries (b) Molecular optimization}}}\par\vspace{1pt}
\setlength{\panelmetricwidth}{\textwidth}
\addtolength{\panelmetricwidth}{-\panelmethodwidth}
\divide\panelmetricwidth by 12
\begin{tabular}{@{}L{\panelmethodwidth}*{12}{C{\panelmetricwidth}}@{}}
\toprule
& \multicolumn{2}{c}{\textsc{QED}} & \multicolumn{2}{c}{\textsc{LogP}} & \multicolumn{2}{c}{\textsc{Solubility}} & \multicolumn{2}{c}{\textsc{DRD2}} & \multicolumn{2}{c}{\textsc{JNK3}} & \multicolumn{2}{c}{\textsc{GSK3B}} \\
\cmidrule(lr){2-3}\cmidrule(lr){4-5}\cmidrule(lr){6-7}\cmidrule(lr){8-9}\cmidrule(lr){10-11}\cmidrule(lr){12-13}
\textbf{Method} &
\twolineheight SR & \twolineheight $\Delta$ &
\twolineheight SR & \twolineheight $\Delta$ &
\twolineheight SR & \twolineheight $\Delta$ &
\twolineheight SR & \twolineheight $\Delta$ &
\twolineheight SR & \twolineheight $\Delta$ &
\twolineheight SR & \twolineheight $\Delta$ \\
\midrule
Student, all-task SFT & 0.68 & 0.153 & 0.58 & 0.349 & 0.58 & 0.446 & 0.46 & 0.030 & 0.30 & -0.008 & 0.41 & 0.026 \\
MIP-routed SFT profile & 0.81 & 0.188 & 0.78 & 0.701 & 0.76 & 0.786 & 0.59 & 0.136 & 0.35 & -0.042 & 0.58 & 0.066 \\
\midrule
Vanilla OPD & 0.82 & 0.184 & 0.74 & 0.632 & 0.60 & 0.490 & 0.41 & 0.036 & 0.27 & -0.036 & 0.41 & \textbf{0.034} \\
Benchmark-family OPD & 0.78 & 0.163 & 0.58 & 0.377 & 0.66 & 0.632 & 0.45 & 0.057 & 0.26 & -0.026 & 0.43 & 0.029 \\
MOPD & 0.76 & 0.167 & 0.70 & 0.641 & 0.63 & 0.506 & 0.50 & 0.032 & 0.18 & -0.032 & 0.42 & 0.020 \\
Homogeneous-RR & 0.74 & 0.164 & 0.64 & 0.557 & \textbf{0.71} & 0.692 & \textbf{0.57} & 0.062 & 0.28 & -0.022 & \textbf{0.46} & 0.023 \\
DanceOPD & \textbf{0.86} & \textbf{0.194} & 0.66 & 0.561 & 0.70 & 0.586 & 0.47 & 0.053 & 0.28 & -0.030 & 0.45 & 0.031 \\
Random-group OPD & 0.78 & 0.165 & \textbf{0.76} & \textbf{0.652} & 0.69 & 0.536 & 0.46 & 0.040 & 0.26 & -0.034 & 0.43 & 0.014 \\
\rowcolor{chemopdblue!20}\textbf{ChemOPD} & \textbf{0.86} & 0.167 & 0.68 & 0.601 & \textbf{0.71} & \textbf{0.718} & 0.55 & \textbf{0.098} & \textbf{0.29} & \textbf{-0.020} & \textbf{0.46} & 0.029 \\
\bottomrule
\end{tabular}

\par\vspace{4pt}
\makebox[\textwidth][c]{{\small\textcolor{tableink}{\bfseries (c) Reaction reasoning}}}\par\vspace{1pt}
\setlength{\panelmetricwidth}{\textwidth}
\addtolength{\panelmetricwidth}{-\panelmethodwidth}
\divide\panelmetricwidth by 11
\begin{tabular}{@{}L{\panelmethodwidth}*{11}{C{\panelmetricwidth}}@{}}
\toprule
& \multicolumn{2}{c}{$\mathrm{Fwd}_{\mathrm{by}}$} & \multicolumn{2}{c}{$\mathrm{Fwd}_{\mathrm{major}}$} & \textsc{MechSel} & \multicolumn{2}{c}{\textsc{NEPP}} & \multicolumn{2}{c}{\textsc{Condition}} & \multicolumn{2}{c}{\textsc{Retro}} \\
\cmidrule(lr){2-3}\cmidrule(lr){4-5}\cmidrule(lr){6-6}\cmidrule(lr){7-8}\cmidrule(lr){9-10}\cmidrule(lr){11-12}
\textbf{Method} &
\twolineheight FTS$\uparrow$ & \twolineheight Top-1 &
\twolineheight FTS$\uparrow$ & \twolineheight Top-1 &
\twolineheight Acc. &
\twolineheight FTS$\uparrow$ & \twolineheight Top-1 &
\twolineheight FTS$\uparrow$ & \twolineheight Top-1 &
\twolineheight FTS$\uparrow$ & \twolineheight Top-1 \\
\midrule
Student, all-task SFT & 0.040 & 0.00 & 0.274 & 0.06 & 0.35 & 0.266 & 0.05 & 0.025 & 0.02 & 0.245 & 0.00 \\
MIP-routed SFT profile & 0.044 & 0.00 & 0.469 & 0.26 & 0.54 & 0.640 & 0.35 & 0.007 & 0.03 & 0.482 & 0.03 \\
\midrule
Vanilla OPD & 0.043 & 0.00 & 0.455 & 0.09 & 0.38 & 0.504 & 0.07 & 0.008 & 0.02 & 0.318 & 0.01 \\
Benchmark-family OPD & 0.046 & 0.00 & 0.411 & 0.15 & 0.42 & 0.506 & 0.09 & 0.000 & 0.00 & 0.437 & 0.02 \\
MOPD & 0.033 & 0.00 & 0.398 & 0.17 & \textbf{0.45} & 0.462 & \textbf{0.13} & \textbf{0.028} & \textbf{0.03} & 0.437 & 0.01 \\
Homogeneous-RR & 0.046 & 0.00 & 0.392 & 0.11 & 0.00 & 0.000 & 0.00 & 0.000 & 0.00 & 0.370 & 0.00 \\
DanceOPD & 0.023 & 0.00 & 0.303 & 0.09 & 0.38 & 0.502 & 0.08 & 0.001 & 0.00 & 0.386 & 0.01 \\
Random-group OPD & 0.040 & 0.00 & 0.439 & 0.17 & 0.37 & 0.505 & 0.06 & 0.000 & 0.00 & 0.433 & 0.00 \\
\rowcolor{chemopdblue!20}\textbf{ChemOPD} & \textbf{0.051} & 0.00 & \textbf{0.465} & \textbf{0.18} & 0.39 & \textbf{0.521} & \textbf{0.13} & 0.023 & \textbf{0.03} & \textbf{0.443} & \textbf{0.03} \\
\bottomrule
\end{tabular}
\endgroup
\end{table*}

\begin{figure*}[ht]
\centering
\includegraphics[width=\textwidth]{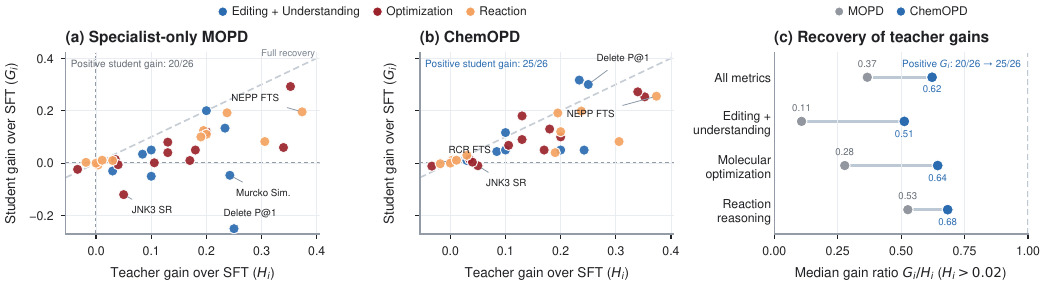}
\caption{Teacher and student gains over the common SFT initializer. For metric
  $i$, $H_i$ is the direction-aligned gain of the routed teacher, and $G_i$ is
  the corresponding student gain; positive values indicate improvement. Panels
  (a--b) plot $G_i$ against $H_i$ for specialist-only MOPD and ChemOPD across all
  31 metrics. Colors denote task families, and the dashed diagonal $G_i=H_i$
  marks full recovery of the teacher gain. Panel (c) reports the median ratio
  $G_i/H_i$ overall and within each task family for the 26 metrics with
  $H_i>0.02$. Gray and blue denote MOPD and ChemOPD, respectively. All gains
  retain their native metric units; metric columns are descriptive and are not
  independent trials.}
\label{fig:opd-headroom-recovery}
\end{figure*}

\noindent \textbf{Relating student gains to teacher headroom.}
Positive teacher headroom does not ensure student improvement: on deletion,
MOPD regresses from the SFT initializer despite its stronger routed teacher,
whereas ChemOPD improves. Headroom recovery summarizes this relationship by
dividing student gain by routed-teacher gain over the common initializer,
reversing error directions; one denotes recovery of the teacher's observed
gain. To avoid near-zero denominators, we summarize the 26 columns with
headroom above 0.02 in native metric units. The reported median rises from
0.37 for MOPD to 0.62 for ChemOPD (Figure~\ref{fig:opd-headroom-recovery}). The largest recovery increases occur in editing and understanding
(0.11 to 0.51) and optimization (0.28 to 0.64). ChemOPD shows its largest recovery gains in families where specialist-only
OPD realizes relatively little of the available teacher headroom.
Recovery describes student improvement relative to the same specialists,
not specialist-specific knowledge transfer; full definitions and family
summaries are in Appendix~\ref{app:headroom}.

\subsection{Ablations and Training Behavior}
\label{sec:ablations}

\noindent \textbf{Specialist weight and warm-up.}
We compare ChemOPD with a no-warm-up variant that uses the same final
specialist weight of 0.50 from the first update. ChemOPD with warm-up matches
or improves 24 of 31 metrics, most clearly in reaction reasoning: retrosynthesis FTS
increases from 0.303 to 0.443. QED property gain instead favors constant
weighting (0.191 versus 0.167). Thus, matching the final specialist weight is insufficient to reproduce the
full integration strategy. The warm-up schedule also controls when and how
specialist guidance is introduced during training.

\begin{figure*}[t]
\centering
\includegraphics[width=\textwidth]{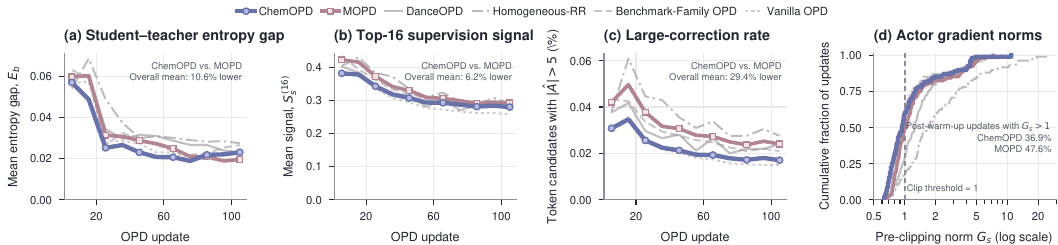}
\caption{OPD training dynamics. Panels (a--c) show 10-update averages of
  (a) the student-teacher entropy gap $E_b$, (b) the signed top-16 supervision signal $S_s^{(16)}$, and (c) the percentage of token candidates
  with $|\hat A|>5$. Panel (d) shows the empirical CDF of the pre-clipping actor
  gradient norm $G_s$; the dashed line marks $G_s=1$. Annotations compare
  ChemOPD with MOPD, which uses the same prompts and specialist routes. See
  Appendix~\ref{app:opd-training-diagnostics} for definitions.}
\label{fig:opd-training-dynamics}
\end{figure*}

\noindent \textbf{Training behavior.}
Figure~\ref{fig:opd-training-dynamics} relates the student results to changes
in supervision and update scale. Compared with MOPD, ChemOPD has a smaller
student-specialist entropy gap, a 29.4\% lower extreme-candidate rate, and
fewer gradient-norm exceedances after learning-rate warm-up. The entropy
channel uses the routed specialist, not a generalist-specialist mixture.
These observations do not by themselves identify the causal role of each
training signal. However, they show that ChemOPD introduces specialist supervision with a more moderate training signal while
maintaining broader capability integration.
Diagnostic definitions are provided in
Appendix~\ref{app:opd-training-diagnostics}.
Checkpoint trajectories provide a complementary view
(Appendix~\ref{app:checkpoint-evaluation}).
MOPD improves the reaction summaries faster early in training,
whereas ChemOPD maintains a higher understanding summary and
finishes ahead on both reaction summaries.
Thus, faster early gains in one task family do not necessarily
yield a stronger final cross-task profile.

\section{Conclusion}
We introduced \method{}, a framework for organizing teacher specialization
and integrating chemical capabilities into a unified student.
SFT-gradient affinities and constrained mixed-integer programming determine
which tasks specialists learn together. Anchor-residual OPD then combines
their guidance with generalist supervision, gradually increasing the
specialist contribution. Experiments on ChemCoTBench show task-dependent
teacher gains from affinity-guided grouping and student improvements over
generalist-only and specialist-only distillation across multiple chemical
tasks. Component ablations favor the full training rule over specialist-only supervision and constant mixing on most reported metrics, although individual tasks can favor the alternatives. Together, these findings support considering
not only how teachers develop specialized strengths, but also how those
strengths complement generalist supervision when training a unified chemical
reasoner.

\clearpage

\bibliography{iclr2027_conference}
\bibliographystyle{iclr2027_conference}

\appendix
\setcounter{topnumber}{5}
\setcounter{bottomnumber}{3}
\setcounter{totalnumber}{8}
\renewcommand{\topfraction}{0.95}
\renewcommand{\bottomfraction}{0.90}
\renewcommand{\textfraction}{0.05}
\renewcommand{\floatpagefraction}{0.85}
\renewcommand{\dbltopfraction}{0.95}
\renewcommand{\dblfloatpagefraction}{0.85}

\section{AI Use Statement}
In preparing this manuscript, generative AI tools were used for language
editing, manuscript restructuring, and consistency checking. All scientific
claims, equations, experimental results, citations, and AI-assisted revisions
were reviewed by the authors. The authors take responsibility for the final
content of the submission.

\section{Benchmark Tasks, Metrics, and Data Protocol}
\label{app:tasks}

\begin{table*}[!htbp]
\caption{20 chemical reasoning subtasks used in
the study. Detailed task semantics and metric calculations are given in
Appendix~\ref{app:task-metrics-protocol}.}
\centering
\small
\begin{tabular}{p{0.16\textwidth}p{0.42\textwidth}p{0.32\textwidth}}
\toprule
Family & Subtasks & Primary evaluation dimensions \\
\midrule
Editing & Add, Delete, Substitution & P@1 for a valid output satisfying the requested functional-group change \\
Understanding & Murcko scaffold, Equivalence, functional-group count, ring count, ring-system recognition & Murcko Morgan-Tanimoto; equivalence/ring-system accuracy; FG/ring MAE \\
Optimization & QED, LogP, solubility, DRD2, JNK3, GSK3B & Property gain $\Delta$ and success rate $\mathrm{SR}(\Delta>0)$; validity/scaffold diagnostics \\
Reaction & Fwd$_{\mathrm{by}}$, Fwd$_{\mathrm{major}}$, MechSel, NEPP, Condition/RCR, Retro & Top-1 chemical match and FTS for SMILES tasks; MechSel accuracy \\
\bottomrule
\end{tabular}
\end{table*}

\subsection{Dataset Construction and Reaction Protocol}
\label{app:data}

We use the released ChemCoTBench training and evaluation protocol for all
subtasks except retrosynthesis, for which we construct a product-string-disjoint
holdout from the released training pool as described below.

\subsection{Task Semantics and Metric Calculations}
\label{app:task-metrics-protocol}

We follow the task taxonomy and structured-output evaluation of
ChemCoTBench, using the 20 tasks in
our protocol. Editing asks the model to return a molecule after adding,
deleting, or substituting a specified functional group. Understanding covers
Murcko-scaffold extraction, SMILES equivalence (whether two strings denote the
same molecule), functional-group and ring counting, and binary ring-system
recognition. Optimization asks for a molecule that improves one requested
property: QED (drug-likeness)~\citep{bickerton2012quantifying}, LogP (lipophilicity), aqueous solubility
(ESOL)~\citep{delaney2004esol}, or a target-specific oracle score (DRD2, JNK3, or GSK3B). Reaction reasoning
contains forward byproduct and major-product prediction, mechanism-route
selection (MechSel), next-elementary-step product prediction (NEPP), reaction
condition recommendation (RCR/Condition), and single-step retrosynthesis
(Retro). The forward views predict products from reactants, NEPP predicts the
next-step product, MechSel selects a reaction route, RCR predicts reaction
conditions, and Retro infers reactants from a product. The split and view counts
used here are defined below. 

For a task with $N$ evaluation examples, let $V$ denote the examples with a
valid or recognized answer. Whenever validity is reported, it is
\begin{equation}
\mathrm{Validity}=\frac{|V|}{N}.
\label{eq:app-validity}
\end{equation}

For categorical subtasks, we report accuracy
\begin{equation}
\mathrm{Acc}=\frac{1}{|S|}\sum_{i\in S}
\mathbf{1}[\hat y_i=y_i],
\label{eq:app-accuracy}
\end{equation}
after the task-specific normalization used by the evaluator. The scored set
$S$ is the recognized-answer subset for SMILES equivalence (a valid Yes/No
answer), but is the full evaluation set (with an unparseable choice counted as
incorrect) for ring-system recognition and multiple-choice MechSel. Editing
$\mathrm{P@1}$ uses the total
evaluation set as its denominator, with a success requiring a parseable output
whose functional-group count changes satisfy the requested Add, Delete, or
Substitution operation; equivalently,
$\mathrm{P@1}=N_{\mathrm{success}}/N$, where $N_{\mathrm{success}}$ is the
number of outputs passing that operation check. For functional-group and ring
counts, we instead report
mean absolute error over parseable outputs,
\begin{equation}
\mathrm{MAE}=\frac{1}{|V|}\sum_{i\in V}|\hat c_i-c_i|,
\label{eq:app-mae}
\end{equation}
and report the valid-output rate separately; thus editing P@1 is a one-output
operation-success rate rather than an exact-SMILES rate. Murcko extraction~\citep{bemis1996properties} is
scored by
the Tanimoto similarity of radius-2 Morgan fingerprints (1,024 bits)
~\citep{rogers2010extended} for the
predicted and reference scaffolds; an identical scaffold therefore scores one.
For binary fingerprints $a,b$, the Tanimoto score is
$T(a,b)=|a\cap b|/|a\cup b|$, where the sets denote active bits.

For molecule optimization, we retain the released ChemCoTBench property
oracles and aggregation rules, while a local ID-aligned wrapper extracts the
generated molecule from each model response. Let $N$ be the total number of
evaluation examples for one property, $p$ the requested property, and $m_i$
and $\hat m_i$ the source and generated molecules. For a successfully scored
example, define the raw improvement
\[\delta_i=p(\hat m_i)-p(m_i).\]
If a prediction is missing, cannot be extracted, is invalid, or cannot be
scored by the property oracle, its improvement is set to zero. Thus the
zero-padded vector $\mathbf{x}=(x_1,\ldots,x_N)$ contains the raw $\delta_i$
values for scorable examples and zeros otherwise. Let
$L=P_5(\mathbf{x})$ and $U=P_{95}(\mathbf{x})$ be the 5th and 95th
percentiles and define $\tilde{x}_i=\operatorname{clip}(x_i,L,U)$. The two
reported optimization quantities are
\begin{equation}
\begin{aligned}
\overline{\Delta}_{\mathrm{display}}
&=\frac{1}{N}\sum_{i=1}^{N}\tilde{x}_i,\\
\mathrm{SR}
&=\frac{1}{N}\sum_{i=1}^{N}\mathbf{1}[x_i>0].
\end{aligned}
\label{eq:app-optimization}
\end{equation}
Accordingly, the displayed mean property change uses the winsorized,
zero-padded improvements, whereas success rate uses the raw zero-padded
improvements with the strict criterion $x_i>0$. Both use all evaluation
examples for that property as the denominator. These rules follow the released
benchmark evaluator; the local wrapper adds robust output extraction, ID
alignment, and per-sample diagnostics without changing the released property
oracles or aggregate definitions. When reported, hard scaffold consistency is
exact equality of the source and generated Murcko scaffolds, and soft
consistency is their scaffold fingerprint Tanimoto. For reaction SMILES
generation, \(\mathrm{Top\mbox{-}1}\) is an
RDKit-parseable chemical match (InChI equality in the evaluator), while the
fingerprint similarity score (FTS) is the validity-gated mean of three
Tanimoto similarities:
\begin{equation}
\mathrm{FTS}=\mathrm{Validity}\,
\frac{T_{\mathrm{RDK}}+T_{\mathrm{MACCS}}+T_{\mathrm{Morgan}}}{3}.
\label{eq:app-fts}
\end{equation}
The three fingerprints compare the generated and reference molecules; the
released evaluator averages their per-example similarities (invalid parses
contribute zero) and gates the result by the aggregate validity rate, where
validity is the fraction of RDKit-parseable predictions. Higher values are
better for all primary metrics except FG/Ring MAE, for which lower values are
better. Ring-Sys Sim denotes the binary
ring-system accuracy described above. These definitions follow the released
benchmark evaluator;
our protocol-specific counts, filtering, and retrosynthesis holdout are given
in the remainder of this section.

\textbf{Overall corpus.}
Each SFT instance pairs a benchmark prompt with its released structured-CoT
assistant target; prompt tokens provide context but are masked from the SFT
loss. Table~\ref{tab:data-counts} gives the resulting task-instance counts. The
evaluation protocol contains 1,595 nominal task views originating from 1,495
source records. The only source-to-multiple-view expansion is forward synthesis:
each of the 100 released forward records is exposed once as a major-product view
and once as a byproduct view. The paired views share the same prompt and source
identifier but have distinct scalar targets and view-suffixed evaluation
identifiers. Nineteen byproduct views have empty released references and are
omitted by the reaction evaluator, leaving 1,576 reference-bearing per-sample
evaluation rows. This number is not a universal denominator for every reported
metric, because task-specific evaluators may condition aggregation on successful
output parsing or other benchmark-defined validity rules. For every
non-retrosynthesis task, we preserve the released evaluation data; the only
split change is the deterministic retrosynthesis holdout described below.

\begin{table}[ht]
\caption{SFT training instances and evaluation task views by benchmark family.}
\label{tab:data-counts}
\centering
\small
\begin{tabular}{lrr}
\toprule
Family & SFT training & Evaluation \\
\midrule
Molecular editing & 4,497 & 100 \\
Molecular understanding & 7,718 & 320 \\
Molecular optimization & 5,587 & 600 \\
Reaction reasoning & 10,371 & 575 \\
\midrule
Total & 28,173 & 1,595 \\
\bottomrule
\end{tabular}
\end{table}
The prompt-only OPD file contains one row per SFT instance. Under the current
1,024-token prompt limit, runtime length filtering retains 28,089 prompt rows.
With a prompt batch size of 256 and incomplete batches dropped, one OPD epoch
contains 109 optimizer batches and samples 27,904 prompts.

\textbf{Normalization of released reaction tasks.}
For non-retrosynthesis reaction tasks, we preserve the released training and
evaluation data. The forward task is represented differently in SFT and
evaluation. For SFT, we retain the separately released structured-CoT corpora as
provided, comprising 2,788 major-product training rows and 890 byproduct
training rows. These corpora are not formed by duplicating a common set of
forward reactions: their identifiers and reaction coverage differ, and their
prompts explicitly request major-product or byproduct prediction, respectively.
Accordingly, the reported SFT counts are training rows rather than counts of
distinct reactions. For evaluation, by contrast, each of the 100 released
forward records contains both targets and is exposed as two scalar-target views
sharing the same prompt, one for the major product and one for the byproduct. Table~\ref{tab:rxn-data-counts}
reports the resulting counts.

\begin{table}[!htbp]
\caption{Reaction-task instance counts under the current protocol. Each
released forward-synthesis example contributes one major-product and one
byproduct view.}
\label{tab:rxn-data-counts}
\centering
\small
\begin{tabular}{lrrr}
\toprule
Subtask & Train & Eval. & Scored \\
\midrule
Byproduct prediction & 890 & 100 & 81 \\
Major-product prediction & 2,788 & 100 & 100 \\
Mechanism selection & 229 & 100 & 100 \\
NEPP & 1,257 & 85 & 85 \\
RCR & 3,142 & 90 & 90 \\
Retrosynthesis & 2,065 & 100 & 100 \\
\midrule
Total & 10,371 & 575 & 556 \\
\bottomrule
\end{tabular}
\end{table}

\textbf{Retrosynthesis test set re-split.}
  Inspection of the released retrosynthesis test snapshot shows that 81 of the
  100 reference targets correspond to multiple reactants whose SMILES strings
  are concatenated without the \texttt{.} component separators required by the
  task output format; the other 19 contain a single reactant. For the
  multi-reactant targets, the molecular boundaries cannot be recovered
  unambiguously from the concatenated strings alone. Heuristically inserting
  separators would therefore modify the ground-truth annotations rather than
  reliably restore them. We construct a new evaluation holdout from the 2,165 records in the
  released retrosynthesis structured-CoT training pool, whose metadata preserves
  the reactant targets. We select 100 records for evaluation and retain the
  remaining 2,065 records for training.

  \textbf{Construction of the product-string-disjoint holdout.}
  For each of the 2,165 source training records, we read
 raw data, collect its dot-separated fragments, strip leading and
  trailing whitespace, discard empty fragments, sort the remaining fragment
  strings lexicographically, and rejoin them with \texttt{.} to obtain a
  normalized product-string key. Records sharing the same key are assigned to
  the same partition.  Evaluation examples are selected only from normalized product-string keys that
 occur once in the source training pool. Consequently, once a record is selected
  for evaluation, its normalized product-string key does not appear in the
  retained training set, thereby preventing train-test overlap at the
  normalized product-string level. Under the 100-record evaluation budget, the deterministic selection first
  maximizes reaction-class coverage among the eligible singleton groups. This
  covers 82 of the 87 source reaction classes, corresponding to all classes with
  at least one singleton candidate. The remaining 18 positions are filled
  according to class-frequency deficits, with at most one additional record per
  class during this stage. 
  This construction guarantees product-string disjointness only under the
  normalization described above, which accounts for fragment ordering and
  surrounding whitespace.
\begin{table}[ht]
\caption{Construction of the retrosynthesis holdout.}
\label{tab:retro-split}
\centering
\small
\begin{tabular}{lr}
\toprule
Statistic & Count \\
\midrule
Source records & 2,165 \\
Normalized product-string groups & 1,818 \\
Source reaction classes & 87 \\
Singleton product-string groups & 1,676 \\
Classes with singleton candidates & 82 \\
Training records & 2,065 \\
Evaluation records & 100 \\
Evaluation product groups / exact queries & 100 / 100 \\
Reaction classes represented in evaluation & 82 \\
\bottomrule
\end{tabular}
\end{table}

 We verify that none of the 100 held-out examples overlap with any training or
  affinity-probe data used in our pipeline by record ID, exact retrosynthesis
  prompt, or normalized product-string key, ruling out train-test leakage
  under these matching criteria.
\FloatBarrier

\section{Implementation and Comparison Protocols}
\label{app:implementation}

Tables~\ref{tab:sft-hparams} and~\ref{tab:opd-hparams} summarize the SFT and OPD configurations used for the completed experiments; affinity-probe and MIP settings are reported with the grouping in Table~\ref{tab:mip-hparams}. We use the same data-shuffling protocol across methods, and evaluate each model with temperature-zero decoding. This setup controls the main sources of run-to-run variation and keeps the comparison conditions consistent across methods. The tables retain the principal settings shared across methods. The following subsections specify the comparisons and method-specific batch construction.

\subsection{Comparison Protocols}
\label{app:comparison-protocols}

\textbf{Teacher construction.}
All-task training fits one teacher to the full task collection, whereas
benchmark-family grouping trains specialists on the benchmark's semantic
families. The MIP construction uses affinity-guided groups with limited
overlap. Human-defined controls combine the understanding and optimization
families, understanding and reaction, or editing and optimization. The
capacity-matched random
construction preserves the group capacity but assigns tasks randomly. Human controls are compared
only on tasks they cover; missing entries are not failures. Complete scores
and the human-control visualization appear in
Appendix~\ref{app:teacher-results}.

\textbf{Student integration.}
Vanilla OPD uses the all-task teacher, and benchmark-family OPD uses its family
specialists. Specialist-only MOPD \citep{ma2026mopd} shares ChemOPD's SFT
student initialization, MIP specialist checkpoints, fixed routes, prompt
stream, and actor-update budget, but omits the anchor. Random-group OPD
retains the ChemOPD objective with a randomly constructed specialist
collection. We use it to compare teacher collections under the same
integration objective.

Homogeneous-RR and the DanceOPD-style control change how routed training units
are assembled into an update. The latter adapts a sampling strategy from
\citet{zhou2026danceopd}, not its flow-matching image-model objective.
The exact route sampling and task exposures are specified below. Equal
update, prompt, and rollout budgets do not make these controls task-exposure
matched to ChemOPD and MOPD.

\subsection{Batch Construction and Teacher Scoring}
\label{app:batch-construction}

\textbf{Method-specific OPD batch construction.}
For ChemOPD and MOPD, runtime prompt-length filtering retains 28,089 of the
28,173 prepared prompt rows. With batch size 256 and drop-last batching, the
standard shuffled stream therefore yields 109 distributed optimizer updates and
27,904 prompt draws. Four rollouts
per prompt produce 111,616 rollout sequences in total. The resulting task-family exposure is
10,222 reaction prompts, 12,138 editing/understanding prompts, and 5,544
optimization prompts. Homogeneous-RR preserves the same total update, prompt,
and rollout budgets, but each complete 256-prompt update comes from a single
specialist pool. The three specialists are visited in strict round-robin order,
yielding 37/36/36 updates and 9,472/9,216/9,216 prompts for
reaction, editing/understanding, and optimization, respectively. 

The DanceOPD baseline also preserves 109 updates and 256 prompts per
update, but partitions each update into four contiguous 64-prompt homogeneous
route slots. Each slot independently samples one route, with replacement, from
the all-task teacher and the three specialists. The four slots are concatenated,
scored by their selected teachers, restored to batch order, and jointly passed
to one actor optimizer update; they are not four separate backward/optimizer
steps.  Thus, both sampling-based controls match ChemOPD/MOPD in the total numbers of
  updates, prompts, and rollouts, but distribute the training prompts differently
  across task families.

The DanceOPD comparison retains only the aspects that transfer naturally
to this language-model setting-on-policy rollouts, hard teacher routing,
route sampling with replacement, and aggregation of several routed units into
one update. It does not reproduce the original flow-matching velocity field,
diffusion trajectory-state queries, timestep sampling, velocity-MSE objective,
LoRA teacher-field composition, or off-policy teacher-trajectory branch.

\begin{table}[ht]
\caption{Core supervised fine-tuning settings. Microbatch size, gradient
accumulation, and gradient checkpointing were adjusted to model size and GPU
allocation; the optimizer and data-processing settings below were shared.}
\label{tab:sft-hparams}
\centering
\begingroup
\small
\renewcommand{\arraystretch}{1.08}
\setlength{\tabcolsep}{4pt}
\begin{tabular}{@{}L{0.35\columnwidth}L{0.60\columnwidth}@{}}
\toprule
\textbf{Hyperparameter} & \textbf{Value} \\
\midrule
Teacher initialization & Qwen3-8B-Base \\
Student initialization & Qwen3-1.7B-Base \\
Fine-tuning scope & Full-parameter \\
Training objective & Assistant-token causal LM loss; prompt masked \\
Epochs & 3 \\
Optimizer & AdamW, $\beta=(0.9,0.999)$, weight decay $0.01$ \\
Learning rate & $2\times10^{-5}$, constant \\
Maximum sequence length & 4096 tokens \\
Numerical precision & bfloat16 \\
Gradient clipping & Global norm $1.0$ \\
Distributed backend & DDP \\
Random seed & 42 \\
\bottomrule
\end{tabular}
\endgroup
\end{table}

\begin{table}[ht]
\caption{OPD hyperparameters shared by the completed methods. The
method-specific batch construction used by Homogeneous-RR and DanceOPD is
described in Appendix~\ref{app:batch-construction}; the implemented distillation
signal and actor objective are defined in Appendix~\ref{app:opd-training-diagnostics}.}
\label{tab:opd-hparams}
\centering
\begingroup
\small
\renewcommand{\arraystretch}{1.08}
\setlength{\tabcolsep}{4pt}
\begin{tabular}{@{}L{0.35\columnwidth}L{0.60\columnwidth}@{}}
\toprule
\textbf{Hyperparameter} & \textbf{Value} \\
\midrule
Student initialization & All-task SFT Qwen3-1.7B \\
Training budget & 109 distributed optimizer updates \\
Prompt batch size & 256 \\
Rollouts per prompt & 4 \\
Actor minibatch & 256 prompts / 1,024 rollout sequences globally \\
Computational microbatching & Dynamic per-GPU packing; 4,096-token budget \\
Prompt/response limit & 1024 / 1024 tokens \\
Optimizer & AdamW, $\beta=(0.9,0.999)$, weight decay $0.01$ \\
Peak learning rate & $1.2\times10^{-5}$ \\
Learning-rate schedule & Cosine; 5\% warm-up; minimum ratio $0.1$ \\
Gradient clipping & Global norm $1.0$ \\
Distillation signal & Detached student-top-16 reverse-KL-style discrepancy \\
Student support weighting & Top-16 by student; $q^S$ renormalized on support \\
Teacher log-probabilities & Full-softmax values gathered on student top-16 \\
Actor objective & On-policy clipped policy-gradient surrogate (PPO-style)\\
Loss aggregation & Candidate sum; masked response-token mean within dynamic microbatches \\
Rollout/teacher temperature & $0.7/0.7$ \\
Rollout truncation & $p=1.0$; no top-$k$ truncation \\
External correctness reward & None \\
Thinking mode & Disabled \\
Data order and seed & Shuffled; seed 42 \\
Reported checkpoint & Step 109 \\
\bottomrule
\end{tabular}
\endgroup
\end{table}

\FloatBarrier

\section{Task Grouping and Construction Settings}
\label{app:groups}

Algorithm~\ref{alg:mip} specifies the profile-enumeration and MILP procedure.
The experiments use $m=3$, group sizes from four to eight,
$q_{\max}=3$, and membership penalty $\lambda=0.1$. The selected $8/8/6$
profile and fixed routes are listed in Table~\ref{tab:mip-groups}. The table provides the complete task assignment and fixed OPD routes; LogP and
solubility are the only shared memberships. Table~\ref{tab:mip-hparams}
collects the probe and solver settings used to obtain this assignment.

\begin{algorithm}[ht]
\caption{Soft-overlap MIP task grouping}
\label{alg:mip}
\footnotesize
\begin{algorithmic}[1]
\Require Cosine affinity $C\in\mathbb{R}^{n\times n}$, group count $m$,
size bounds, membership bounds, overlap penalty $\lambda$
\Ensure Groups $G_1,\ldots,G_m$
\State $S\gets\Call{ZScoreOffDiagonal}{C}$
\State $\mathcal{K}\gets\Call{FeasibleProfiles}{n,m,k_{\min},k_{\max},q_{\max}}$
\State $(z^\star,x^\star)\gets(-\infty,\varnothing)$
\For{$(k_1,\ldots,k_m)\in\mathcal{K}$}
    \State Create binary memberships $x_{ig}$ and pair variables $y_{ijg}$
    \State Add exact group-size and task-coverage constraints
    \State Add $y_{ijg}=x_{ig}x_{jg}$ through standard linear constraints
    \State $z\gets\sum_g k_g^{-1}\sum_{i<j}W_{ij}y_{ijg}
    -\lambda\sum_{i,g}x_{ig}$
    \State $(\hat z,\hat x)\gets\Call{SolveMILP}{\max z}$
    \If{$\hat x$ is feasible and $\hat z>z^\star$}
        \State $(z^\star,x^\star)\gets(\hat z,\hat x)$
    \EndIf
\EndFor
\State \Return $G_g=\{i:x^\star_{ig}=1\}$ for all $g$
\end{algorithmic}
\end{algorithm}

\begin{table*}[ht]
\caption{Affinity-guided soft-overlap groups used by the current reaction
protocol experiments. LogP and Solubility occur in both Teacher 1 and Teacher
3; their fixed OPD route is Teacher 3.}
\label{tab:mip-groups}
\centering
\small
\begin{tabular}{p{0.09\textwidth}p{0.57\textwidth}rp{0.17\textwidth}}
\toprule
Teacher & SFT subtasks & Rows & Fixed OPD route \\
\midrule
Teacher 1 & Opt/LogP, Opt/Solubility, Reaction/Byproduct, Reaction/Major product, Reaction/Mechanism selection, Reaction/NEPP, Reaction/RCR, Reaction/Retrosynthesis & 13,177 & All Reaction tasks \\
Teacher 2 & Edit/Add, Edit/Delete, Edit/Substitution, Understanding/Murcko scaffold, Understanding/Equivalence, Understanding/Functional-group count, Understanding/Ring count, Understanding/Ring-system scaffold & 12,215 & Editing and Understanding \\
Teacher 3 & Opt/DRD2, Opt/GSK3B, Opt/JNK3, Opt/LogP, Opt/QED, Opt/Solubility & 5,587 & All Optimization tasks \\
\bottomrule
\end{tabular}
\end{table*}

\begin{table}[!htbp]
\caption{Affinity-probe and MIP settings used to construct the three ChemOPD
specialists. The selected profile is a practical solver output rather than a
claim of a unique globally optimal grouping.}
\label{tab:mip-hparams}
\centering
\begingroup
\small
\renewcommand{\arraystretch}{1.08}
\setlength{\tabcolsep}{4pt}
\begin{tabular}{@{}L{0.35\columnwidth}L{0.60\columnwidth}@{}}
\toprule
\textbf{Hyperparameter} & \textbf{Value} \\
\midrule
Probe backbone & Qwen3-8B-Base \\
Probe examples per task & 100 train / 50 validation \\
Probe batch/sequence length & 1 / 4096 tokens \\
Gradient representation & Last 2 transformer layers; LM head excluded \\
CountSketch dimension & 8192 \\
Probe precision and seed & bfloat16; 42 \\
Affinity and preprocessing & Cosine; off-diagonal $z$-score \\
Number of task groups & 3 \\
Group-size bounds & 4--8 tasks \\
Membership bounds & 1--3 groups per task \\
Extra-membership bounds & Minimum 0; maximum set by capacity \\
Overlap penalty $\lambda$ & $0.1$ per task--group membership \\
Objective normalization & Group size, $1/|G_g|$ \\
Symmetric compatibility & $W_{ij}=S_{ij}+S_{ji}$ \\
Solver & SciPy/HiGHS MILP \\
Time/gap limits & 300 s / relative gap $0.01$ \\
Selected group sizes & $8/8/6$ (two overlapping tasks) \\
Returned solve statistics & 222.9 s; relative gap $0.00173$ \\
\bottomrule
\end{tabular}
\endgroup
\end{table}

\begin{figure*}[ht]
\centering
\includegraphics[width=\textwidth]{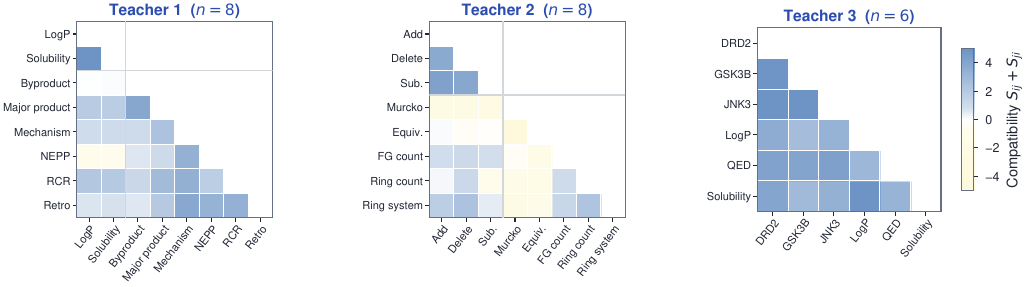}\vspace{-3.9mm}
 \caption{Pairwise compatibility within the three MIP-selected teacher groups.
  Each panel shows $W_{ij}=S_{ij}+S_{ji}$ for every unique off-diagonal task pair
  within one group. Positive and negative values indicate above- and
  below-average compatibility, respectively.}
\label{fig:mip-affinity-blocks}
\end{figure*}

\textbf{Pairwise compatibility within the selected groups.}
  Figure~\ref{fig:mip-affinity-blocks} expands the task-group averages in
  Figure~\ref{fig:mip-compatibility} into the pairwise compatibility scores
  $W_{ij}$ within each selected teacher group. This view shows the variation
  among task pairs within the same group. The complete membership constraints
  and solver settings are reported above.

\FloatBarrier

\section{Full Teacher-Construction Results}
\label{app:teacher-results}

 Table~\ref{tab:teacher-results-full} reports the complete task-level results
  for the teacher constructions compared in
  Section~\ref{sec:analysis-grouping}. For the MIP and benchmark-family
  groupings, each task is evaluated using its predefined teacher route.  Human-defined controls are evaluated only on the tasks covered by
  their respective groups, and blank entries indicate uncovered tasks rather
  than evaluation failures.

\begin{table}[ht]
  \caption{Full task-level results for the teacher constructions. The MIP and
  benchmark-family profiles use predefined task-to-teacher routes. The shaded row
  shows the routed profile of the MIP specialists used by ChemOPD. Blank entries
  indicate tasks not covered by the corresponding human-defined controls.}
\label{tab:teacher-results-full}
\centering
\begingroup
\scriptsize
\renewcommand{\arraystretch}{1.00}
\setlength{\tabcolsep}{0pt}
\setlength{\panelconfigwidth}{4.00cm}
\setlength{\panelmodelwidth}{1.65cm}
\setlength{\panelmethodwidth}{\panelconfigwidth}
\addtolength{\panelmethodwidth}{\panelmodelwidth}

\makebox[\textwidth][c]{{\small\textcolor{tableink}{\bfseries (a) Molecule editing \& understanding}}}\par\vspace{1pt}
\setlength{\panelmetricwidth}{\textwidth}
\addtolength{\panelmetricwidth}{-\panelmethodwidth}
\divide\panelmetricwidth by 8
\begin{tabular}{@{}L{\panelconfigwidth}C{\panelmodelwidth}*{8}{C{\panelmetricwidth}}@{}}
\toprule
& & \multicolumn{3}{c}{\textsc{Editing}} & \multicolumn{5}{c}{\textsc{Understanding}} \\
\cmidrule(lr){3-5}\cmidrule(lr){6-10}
\textbf{Configuration} & \textbf{Model} &
\shortstack[c]{Add\\P@1} &
\shortstack[c]{Delete\\P@1} &
\shortstack[c]{Sub.\\P@1} &
\shortstack[c]{Murcko\\Sim.} &
\shortstack[c]{Equiv.\\Acc.} &
\shortstack[c]{FG\\MAE $\downarrow$} &
\shortstack[c]{Ring\\MAE $\downarrow$} &
\shortstack[c]{Ring-Sys\\Sim.} \\
\midrule
Student, raw & \mbox{Qwen3-1.7B} & 0.15 & 0.00 & 0.00 & 0.59 & 0.49 & 0.15 & 1.05 & 0.67 \\
Teacher, raw & \mbox{Qwen3-8B} & 0.35 & 0.55 & 0.08 & 0.10 & 0.60 & 0.16 & 0.31 & 0.67 \\
\addlinespace[2pt]
Student, all-task SFT & \mbox{Qwen3-1.7B} & 0.30 & 0.45 & 0.38 & 0.53 & 0.58 & 0.11 & 0.20 & 0.72 \\
Teacher, all-task SFT & \mbox{Qwen3-8B} & 0.65 & \textbf{0.70} & 0.52 & 0.74 & \textbf{0.70} & \textbf{0.08} & \textbf{0.10} & 0.82 \\
Benchmark-family SFT & \mbox{Qwen3-8B} & \textbf{0.75} & 0.65 & 0.38 & 0.64 & 0.65 & 0.14 & 0.15 & 0.80 \\
Random-group SFT & \mbox{Qwen3-8B} & \textbf{0.75} & 0.40 & 0.55 & 0.74 & 0.65 & 0.11 & \textbf{0.10} & \textbf{0.85} \\
\rowcolor{chemopdblue!20}\textbf{MIP-routed SFT} & \mbox{Qwen3-8B} & 0.50 & \textbf{0.70} & \textbf{0.62} & \textbf{0.77} & 0.66 & \textbf{0.08} & \textbf{0.10} & 0.82 \\
\addlinespace[2pt]
Human U+O SFT & \mbox{Qwen3-8B} & -- & -- & -- & 0.77 & 0.66 & 0.11 & \textbf{0.10} & 0.78 \\
Human U+R SFT & \mbox{Qwen3-8B} & -- & -- & -- & 0.76 & 0.66 & 0.12 & \textbf{0.10} & 0.75 \\
Human E+O SFT & \mbox{Qwen3-8B} & 0.65 & 0.60 & 0.40 & -- & -- & -- & -- & -- \\
\bottomrule
\end{tabular}

\par\vspace{2pt}
\makebox[\textwidth][c]{{\small\textcolor{tableink}{\bfseries (b) Molecular optimization}}}\par\vspace{1pt}
\setlength{\panelmetricwidth}{\textwidth}
\addtolength{\panelmetricwidth}{-\panelmethodwidth}
\divide\panelmetricwidth by 12
\begin{tabular}{@{}L{\panelconfigwidth}C{\panelmodelwidth}*{12}{C{\panelmetricwidth}}@{}}
\toprule
& & \multicolumn{2}{c}{\textsc{QED}} & \multicolumn{2}{c}{\textsc{LogP}} & \multicolumn{2}{c}{\textsc{Solubility}} & \multicolumn{2}{c}{\textsc{DRD2}} & \multicolumn{2}{c}{\textsc{JNK3}} & \multicolumn{2}{c}{\textsc{GSK3B}} \\
\cmidrule(lr){3-4}\cmidrule(lr){5-6}\cmidrule(lr){7-8}\cmidrule(lr){9-10}\cmidrule(lr){11-12}\cmidrule(lr){13-14}
\textbf{Configuration} & \textbf{Model} &
\twolineheight SR & \twolineheight $\Delta$ &
\twolineheight SR & \twolineheight $\Delta$ &
\twolineheight SR & \twolineheight $\Delta$ &
\twolineheight SR & \twolineheight $\Delta$ &
\twolineheight SR & \twolineheight $\Delta$ &
\twolineheight SR & \twolineheight $\Delta$ \\
\midrule
Student, raw & \mbox{Qwen3-1.7B} & 0.14 & 0.020 & 0.10 & 0.048 & 0.02 & 0.000 & 0.00 & 0.000 & 0.01 & \textbf{0.000} & 0.06 & 0.001 \\
Teacher, raw & \mbox{Qwen3-8B} & 0.17 & 0.025 & 0.14 & 0.101 & 0.34 & 0.211 & 0.12 & 0.016 & 0.11 & -0.002 & 0.09 & 0.001 \\
\addlinespace[2pt]
Student, all-task SFT & \mbox{Qwen3-1.7B} & 0.68 & 0.153 & 0.58 & 0.349 & 0.58 & 0.446 & 0.46 & 0.030 & 0.30 & -0.008 & 0.41 & 0.026 \\
Teacher, all-task SFT & \mbox{Qwen3-8B} & \textbf{0.81} & 0.178 & 0.73 & 0.692 & 0.74 & 0.768 & 0.58 & 0.134 & 0.27 & -0.025 & 0.51 & 0.063 \\
\addlinespace[2pt]
Benchmark-family SFT & \mbox{Qwen3-8B} & \textbf{0.81} & 0.188 & 0.78 & \textbf{0.701} & \textbf{0.76} & \textbf{0.786} & \textbf{0.59} & \textbf{0.136} & \textbf{0.35} & -0.042 & \textbf{0.58} & \textbf{0.066} \\
Random-group SFT & \mbox{Qwen3-8B} & 0.76 & \textbf{0.190} & \textbf{0.79} & 0.663 & 0.72 & 0.553 & 0.52 & 0.085 & 0.28 & -0.023 & 0.51 & 0.062 \\
\rowcolor{chemopdblue!20}\textbf{MIP-routed SFT} & \mbox{Qwen3-8B} & \textbf{0.81} & 0.188 & 0.78 & \textbf{0.701} & \textbf{0.76} & \textbf{0.786} & \textbf{0.59} & \textbf{0.136} & \textbf{0.35} & -0.042 & \textbf{0.58} & \textbf{0.066} \\
\addlinespace[2pt]
Human U+O SFT & \mbox{Qwen3-8B} & 0.76 & 0.185 & 0.78 & 0.672 & \textbf{0.76} & 0.712 & 0.57 & 0.114 & 0.30 & -0.025 & 0.49 & 0.041 \\
Human E+O SFT & \mbox{Qwen3-8B} & 0.75 & 0.190 & 0.77 & 0.645 & 0.67 & 0.693 & 0.51 & 0.128 & 0.25 & -0.022 & 0.57 & 0.065 \\
\bottomrule
\end{tabular}

\par\vspace{2pt}
\makebox[\textwidth][c]{{\small\textcolor{tableink}{\bfseries (c) Reaction reasoning}}}\par\vspace{1pt}
\setlength{\panelmetricwidth}{\textwidth}
\addtolength{\panelmetricwidth}{-\panelmethodwidth}
\divide\panelmetricwidth by 11
\begin{tabular}{@{}L{\panelconfigwidth}C{\panelmodelwidth}*{11}{C{\panelmetricwidth}}@{}}
\toprule
& & \multicolumn{2}{c}{$\mathrm{Fwd}_{\mathrm{by}}$} & \multicolumn{2}{c}{$\mathrm{Fwd}_{\mathrm{major}}$} & \textsc{MechSel} & \multicolumn{2}{c}{\textsc{NEPP}} & \multicolumn{2}{c}{\textsc{Condition}} & \multicolumn{2}{c}{\textsc{Retro}} \\
\cmidrule(lr){3-4}\cmidrule(lr){5-6}\cmidrule(lr){7-7}\cmidrule(lr){8-9}\cmidrule(lr){10-11}\cmidrule(lr){12-13}
\textbf{Configuration} & \textbf{Model} &
\twolineheight FTS$\uparrow$ & \twolineheight Top-1 &
\twolineheight FTS$\uparrow$ & \twolineheight Top-1 &
\twolineheight Acc. &
\twolineheight FTS$\uparrow$ & \twolineheight Top-1 &
\twolineheight FTS$\uparrow$ & \twolineheight Top-1 &
\twolineheight FTS$\uparrow$ & \twolineheight Top-1 \\
\midrule
Student, raw & \mbox{Qwen3-1.7B} & 0.021 & 0.00 & 0.189 & 0.00 & 0.00 & 0.000 & 0.00 & \textbf{0.072} & 0.02 & 0.064 & 0.00 \\
Teacher, raw & \mbox{Qwen3-8B} & 0.043 & 0.00 & 0.437 & 0.00 & 0.04 & 0.000 & 0.00 & 0.042 & \textbf{0.03} & 0.282 & 0.00 \\
\addlinespace[2pt]
Student, all-task SFT & \mbox{Qwen3-1.7B} & 0.040 & 0.00 & 0.274 & 0.06 & 0.35 & 0.266 & 0.05 & 0.025 & 0.02 & 0.245 & 0.00 \\
Teacher, all-task SFT & \mbox{Qwen3-8B} & 0.046 & 0.00 & 0.435 & 0.19 & 0.46 & 0.486 & 0.31 & 0.022 & 0.02 & 0.358 & 0.01 \\
\addlinespace[2pt]
Benchmark-family SFT & \mbox{Qwen3-8B} & 0.037 & 0.00 & 0.451 & 0.21 & \textbf{0.57} & 0.568 & \textbf{0.38} & 0.005 & 0.00 & 0.428 & \textbf{0.03} \\
Random-group SFT & \mbox{Qwen3-8B} & \textbf{0.054} & 0.00 & 0.385 & 0.18 & \textbf{0.57} & 0.458 & 0.26 & 0.034 & 0.00 & 0.405 & 0.02 \\
\rowcolor{chemopdblue!20}\textbf{MIP-routed SFT} & \mbox{Qwen3-8B} & 0.044 & 0.00 & \textbf{0.469} & \textbf{0.26} & 0.54 & \textbf{0.640} & 0.35 & 0.007 & \textbf{0.03} & \textbf{0.482} & \textbf{0.03} \\
\addlinespace[2pt]
Human U+R SFT & \mbox{Qwen3-8B} & 0.024 & 0.00 & 0.419 & 0.22 & 0.50 & 0.625 & 0.32 & 0.071 & 0.02 & 0.419 & 0.02 \\
\bottomrule
\end{tabular}
\endgroup
\end{table}

\begin{figure*}[ht]
\centering
\includegraphics[width=\textwidth]{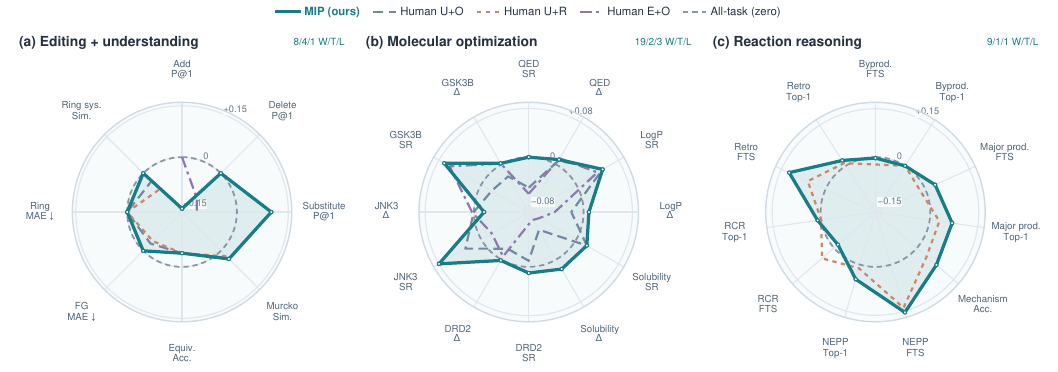}

\caption{Direction-aligned teacher gains from
  Table~\ref{tab:teacher-results-full}. The dashed ring is the generalist
  teacher; outward is better, with MAE signs reversed. Human-control curves
  remain open for uncovered tasks, while MIP-routed follows its fixed route.
  Teal labels report pooled pairwise comparisons between MIP-routed and all
  available human-control results. Each
  metric contributes one comparison for every human control that covers it.
  Metrics retain native units and panel-specific scales.}
\label{fig:human-intuitive-grouping}
\end{figure*}

\textbf{Human-defined grouping controls.}
The Understanding+Optimization, Understanding+Reaction, and
Editing+Optimization controls test whether intuitive family combinations
produce comparable specialists. They often strengthen one capability while
weakening another. For example, Understanding+Reaction improves several
reaction metrics relative to the all-task teacher but gives up understanding
performance. MIP matches or improves on most covered human-control metrics,
although the advantages are not uniform (Figure~\ref{fig:human-intuitive-grouping}).
The curves preserve native metric scales and leave uncovered axes open.

\FloatBarrier

 \section{Robustness to Affinity-Probe Resampling}
  \label{app:affinity-robustness}

  Section~\ref{sec:analysis-grouping} evaluates whether affinity-guided grouping
  produces useful specialist teachers. Here, we study a complementary question:
  Does the estimated affinity structure remain stable when the probe examples
  change? We also test whether these changes lead the MIP to select different
  task groups. This analysis checks whether the grouping reported in the main
  text depends on one particular probe sample.

  \textbf{Resampling protocol.} We use the affinity matrix and its $8/8/6$ grouping as the reference.
  We denote this reference run by $b=0$. We then create $B=10$ new probe draws,
  indexed by $b\in\{1,\ldots,B\}$, from our training split. For each task in each draw, we sample 100 training-probe examples and 50
  validation-probe examples without replacement. Within a draw, the two probe
  subsets are disjoint by normalized prompt. For retrosynthesis, they are also
  disjoint by normalized product-string key. Examples may be reused across
  different draws.

  We keep the Qwen3-8B-Base checkpoint, probed gradient layers, and CountSketch
  map fixed. We also keep the MIP objective and constraints fixed. All ten audit
  runs use a 600-second solver limit and a relative gap of $0.001$. This setting
  is tighter than the solver setting used for the original grouping and reduces
  solver noise in the audit. Across the ten audit runs, only the probe examples
  change. We do not retrain any specialist teacher or student model.

  \textbf{Affinity-stability metrics.}
  For run $b$, let $C^{(b)}$ be the directed cosine-affinity matrix defined in
  Equation~\ref{eq:task-affinity}. Let $S^{(b)}$ be the off-diagonal
  $z$-score of $C^{(b)}$, computed in the same way as in the main grouping
  method. We define the symmetric task-compatibility matrix as
  \begin{equation}
  W^{(b)}=S^{(b)}+S^{(b)\top}.
  \label{eq:affinity-resample-compatibility}
  \end{equation}

  Let $w^{(b)}=\operatorname{vec}_{\triangle}(W^{(b)})$ contain the
  upper-triangular entries of $W^{(b)}$. We have 20 tasks, so
  $w^{(b)}$ contains $\binom{20}{2}=190$ task-pair values. We compare each
  resampled vector with the reference vector using
  \begin{equation}
  \begin{aligned}
  \rho_{\mathrm{rank}}^{(b)}
  &=\operatorname{Spearman}\!\left(w^{(0)},w^{(b)}\right),\\
  \rho_{\mathrm{value}}^{(b)}
  &=\operatorname{Pearson}\!\left(w^{(0)},w^{(b)}\right).
  \end{aligned}
  \label{eq:affinity-resample-correlation}
  \end{equation}
  The Spearman correlation measures whether the task-pair ordering is preserved.
  The Pearson correlation measures whether the compatibility values are
  preserved.

  We next measure local stability. For task $i$, let $N_3^{(b)}(i)$ be the set
  of its three highest-scoring neighbors under $W^{(b)}$. Let
  $J_i^{(b)}$ compare this set with the reference neighbors. We average this
  score over the $n=20$ tasks and denote the result by $\bar J_3^{(b)}$. We also examine task pairs with large reference compatibility. Let
  $\mathcal E_s$ contain the 48 pairs in the top quartile of
  $\lvert W_{ij}^{(0)}\rvert$. We denote their sign agreement by
  $A_{\mathrm{sign}}^{(b)}$. These metrics are
  \begin{equation}
  \begin{aligned}
  J_i^{(b)}
  &=
  \frac{
  \left|N_3^{(0)}(i)\cap N_3^{(b)}(i)\right|
  }{
  \left|N_3^{(0)}(i)\cup N_3^{(b)}(i)\right|
  },\\
  \bar J_3^{(b)}
  &=\frac{1}{n}\sum_{i=1}^{n}J_i^{(b)},\\
  A_{\mathrm{sign}}^{(b)}
  &=
  \frac{1}{|\mathcal E_s|}
  \sum_{(i,j)\in\mathcal E_s}
  \mathbf{1}\!\left[
  \operatorname{sgn}W_{ij}^{(0)}
  =
  \operatorname{sgn}W_{ij}^{(b)}
  \right].
  \end{aligned}
  \label{eq:affinity-neighbor-jaccard}
  \end{equation}
  The sign in $A_{\mathrm{sign}}^{(b)}$ is the sign of the standardized
  compatibility $W_{ij}^{(b)}$. It is not the sign of the raw gradient dot
  product.

  \textbf{Grouping-stability metrics.}
  Let $x^{(0)}$ be the reference MIP assignment and let $x^{(b)}$ be the
  assignment obtained from draw $b$. As in Equation~\ref{eq:mip},
  $x_{ig}=1$ means that task $i$ belongs to group $g$. Teacher labels can be permuted without changing the grouping. We therefore
  compare task co-membership instead of teacher labels. Let $E(x)$ be the set of
  task pairs that share at least one group under assignment $x$. We measure
  grouping agreement using
  \begin{equation}
  J_{\mathrm{group}}^{(b)}
  =
  \frac{
  \left|E(x^{(0)})\cap E(x^{(b)})\right|
  }{
  \left|E(x^{(0)})\cup E(x^{(b)})\right|
  }.
  \label{eq:mip-group-jaccard}
  \end{equation}

  Let $F_b(x)$ be the MIP objective in Equation~\ref{eq:mip}, evaluated using
  $S^{(b)}$. We define the objective gap between the re-solved assignment and
  the reference assignment as
  \begin{equation}
  \Delta F_b
  =
  F_b\!\left(x^{(b)}\right)
  -
  F_b\!\left(x^{(0)}\right).
  \label{eq:mip-objective-gap}
  \end{equation}
  A zero gap means that re-solving the MIP does not improve on the reference
  assignment for that probe draw. Finally, we compare the reference grouping with matched random assignments.
  For each draw, we randomly permute the task labels in $x^{(0)}$ 10,000 times.
  This preserves the $8/8/6$ group sizes and the overlap structure. Let
  $\mu_b^{\mathrm{rand}}$ and $\sigma_b^{\mathrm{rand}}$ be the mean and standard
  deviation of the resulting objective values. We report
  \begin{equation}
  Z_b
  =
  \frac{
  F_b\!\left(x^{(0)}\right)-\mu_b^{\mathrm{rand}}
  }{
  \sigma_b^{\mathrm{rand}}
  }.
  \label{eq:mip-random-zscore}
  \end{equation}
  Thus, $Z_b$ measures how far the reference grouping lies above the matched
  random baseline.

  \begin{figure*}[ht]
  \centering
  \includegraphics[width=\textwidth]{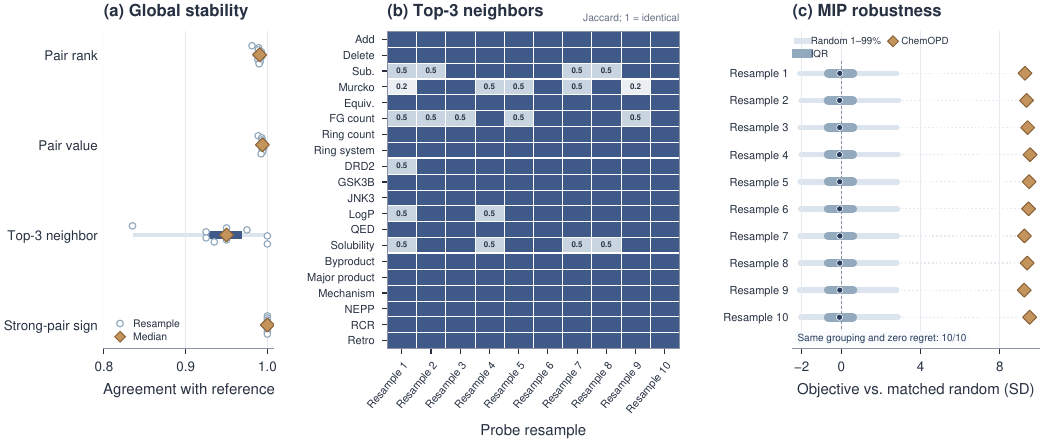}
  \caption{Stability of affinity estimation and MIP grouping under probe
  resampling. (a) Agreement between each resampled affinity matrix and the
  reference. Open circles show the ten draws. Light bars show the full range,
  dark bars show the interquartile range, and diamonds show the median. The
  top-3 neighbor score is averaged over tasks. (b) Task-level top-3 neighbor
  Jaccard scores. A value of one means that all three neighbors are unchanged.
  (c) The objective score of the fixed ChemOPD grouping compared with 10,000
  matched random assignments for each draw. Light and dark bars show the
  1st--99th percentile range and interquartile range of the random scores.
  Diamonds show the fixed ChemOPD grouping.}
  \label{fig:affinity-mip-stability-audit}
  \end{figure*}

  \paragraph{Results.}
  Figure~\ref{fig:affinity-mip-stability-audit}(a) shows strong agreement between
  the resampled affinity matrices and the reference. The median Spearman
  correlation is $0.991$, with a range of $0.981$--$0.994$. The median Pearson
  correlation is $0.994$, with a range of $0.989$--$0.995$. Thus, both the
  ordering and the values of the 190 task-pair compatibilities remain stable. The median value of $\bar J_3^{(b)}$ is $0.950$, with a range of
  $0.835$--$1.000$. Figure~\ref{fig:affinity-mip-stability-audit}(b) shows that
  most tasks keep the same three highest-scoring neighbors. A small number of
  tasks change one or more neighbors in some draws. However, all 48
  high-magnitude reference pairs keep the same sign in every draw. The global
  affinity structure is therefore stable, although a few local rankings change. These local changes do not alter the selected task groups. All ten MIP runs
  recover the reference $8/8/6$ grouping up to a permutation of teacher labels.
  LogP and solubility remain the two overlapping tasks in every run. Therefore,
  $J_{\mathrm{group}}^{(b)}=1$ and $\Delta F_b=0$ for all ten draws. Figure~\ref{fig:affinity-mip-stability-audit}(c) compares the fixed ChemOPD
  grouping with matched random assignments. The fixed grouping scores above all
  10,000 random assignments in every draw. Its median standardized score is
  $9.39$, with a range of $9.23$--$9.52$. The selected grouping therefore
  remains well aligned with each resampled affinity matrix.
  
  This experiment gives two results. First, the estimated task affinities are
  stable across different probe samples. Second, the MIP selects the same task
  groups in all ten runs, and these groups score far above matched random
  assignments. Together, the teacher-performance results in
  Section~\ref{sec:analysis-grouping} and this resampling analysis provide
  evidence that affinity-guided grouping yields task-specific teacher gains and
  remains stable when the probe examples change.

\section{ChemOPD Ablations}
\label{app:chemopd-ablation}

This section provides additional ablation analysis for ChemOPD. We compare the full method with variants that remove the anchor or the warm-up schedule. 

\textbf{Anchor and specialist-weight schedule.}
  Table~\ref{tab:chemopd-component-ablation} compares full ChemOPD with
  Specialist-only (MOPD) and no warm-up under the same initialization,
  specialists, routes, prompts, and update budget. Specialist-only removes the
  anchor, while no warm-up sets $\alpha_g(s)=0.50$ from the first update. Full
  ChemOPD instead increases it from 0 to 0.50 during the first 65\% of training.
  Against Specialist-only, full ChemOPD records 22 wins, four ties, and five
  losses across 31 metrics, including an increase in Solubility property gain
  from $0.506$ to $0.718$. Against no warm-up, it records 23 wins, one tie, and
  seven losses. The main benefit appears in reaction reasoning, with nine wins
  among ten non-tied comparisons and an increase in Retrosynthesis FTS from
  $0.303$ to $0.443$. Together, these comparisons support combining the generalist anchor with a gradual specialist schedule: full ChemOPD improves more reported metrics than either specialist-only supervision or the no-warm-up variant.

\begin{table*}[ht]
\caption{Component ablations for ChemOPD. Each panel follows the native metrics
of Table~\ref{tab:detailed-opd-results}. FG and Ring are lower-is-better.}
\label{tab:chemopd-component-ablation}
\centering
\begingroup
\footnotesize
\renewcommand{\arraystretch}{1.08}
\setlength{\tabcolsep}{0pt}

\makebox[\textwidth][c]{{\small\textcolor{tableink}{\bfseries (a) Molecule editing \& understanding}}}\par\vspace{1pt}
\setlength{\panelmethodwidth}{3.50cm}
\setlength{\panelmetricwidth}{\textwidth}
\addtolength{\panelmetricwidth}{-\panelmethodwidth}
\divide\panelmetricwidth by 8
\begin{tabular}{@{}L{\panelmethodwidth}*{8}{C{\panelmetricwidth}}@{}}
\toprule
& \multicolumn{3}{c}{\textsc{Editing}} & \multicolumn{5}{c}{\textsc{Understanding}} \\
\cmidrule(lr){2-4}\cmidrule(lr){5-9}
\textbf{Method} &
\shortstack[c]{Add\\P@1} &
\shortstack[c]{Delete\\P@1} &
\shortstack[c]{Sub.\\P@1} &
\shortstack[c]{Murcko\\Sim.} &
\shortstack[c]{Equiv.\\Acc.} &
\shortstack[c]{FG\\MAE $\downarrow$} &
\shortstack[c]{Ring\\MAE $\downarrow$} &
\shortstack[c]{Ring-Sys\\Sim.} \\
\midrule
ChemOPD, Specialist-only & \textbf{0.50} & 0.20 & 0.52 & 0.48 & 0.61 & 0.14 & 0.25 & 0.77 \\
ChemOPD, no warm-up & 0.45 & \textbf{0.85} & 0.60 & 0.50 & 0.57 & \textbf{0.09} & 0.20 & 0.73 \\
\rowcolor{chemopdblue!20}\textbf{ChemOPD} & 0.35 & 0.75 & \textbf{0.70} & \textbf{0.58} & \textbf{0.62} & 0.10 & \textbf{0.15} & \textbf{0.83} \\
\bottomrule
\end{tabular}

\par\vspace{4pt}
\makebox[\textwidth][c]{{\small\textcolor{tableink}{\bfseries (b) Molecular optimization}}}\par\vspace{1pt}
\setlength{\panelmetricwidth}{\textwidth}
\addtolength{\panelmetricwidth}{-\panelmethodwidth}
\divide\panelmetricwidth by 12
\begin{tabular}{@{}L{\panelmethodwidth}*{12}{C{\panelmetricwidth}}@{}}
\toprule
& \multicolumn{2}{c}{\textsc{QED}} & \multicolumn{2}{c}{\textsc{LogP}} & \multicolumn{2}{c}{\textsc{Solubility}} & \multicolumn{2}{c}{\textsc{DRD2}} & \multicolumn{2}{c}{\textsc{JNK3}} & \multicolumn{2}{c}{\textsc{GSK3B}} \\
\cmidrule(lr){2-3}\cmidrule(lr){4-5}\cmidrule(lr){6-7}\cmidrule(lr){8-9}\cmidrule(lr){10-11}\cmidrule(lr){12-13}
\textbf{Method} &
\twolineheight SR & \twolineheight $\Delta$ &
\twolineheight SR & \twolineheight $\Delta$ &
\twolineheight SR & \twolineheight $\Delta$ &
\twolineheight SR & \twolineheight $\Delta$ &
\twolineheight SR & \twolineheight $\Delta$ &
\twolineheight SR & \twolineheight $\Delta$ \\
\midrule
ChemOPD, Specialist-only & 0.76 & 0.167 & \textbf{0.70} & \textbf{0.641} & 0.63 & 0.506 & 0.50 & 0.032 & 0.18 & -0.032 & 0.42 & 0.020 \\
ChemOPD, no warm-up & 0.84 & \textbf{0.191} & 0.63 & 0.500 & 0.61 & 0.427 & 0.45 & 0.055 & \textbf{0.32} & \textbf{-0.011} & 0.44 & 0.016 \\
\rowcolor{chemopdblue!20}\textbf{ChemOPD} & \textbf{0.86} & 0.167 & 0.68 & 0.601 & \textbf{0.71} & \textbf{0.718} & \textbf{0.55} & \textbf{0.098} & 0.29 & -0.020 & \textbf{0.46} & \textbf{0.029} \\
\bottomrule
\end{tabular}

\par\vspace{4pt}
\makebox[\textwidth][c]{{\small\textcolor{tableink}{\bfseries (c) Reaction reasoning}}}\par\vspace{1pt}
\setlength{\panelmetricwidth}{\textwidth}
\addtolength{\panelmetricwidth}{-\panelmethodwidth}
\divide\panelmetricwidth by 11
\begin{tabular}{@{}L{\panelmethodwidth}*{11}{C{\panelmetricwidth}}@{}}
\toprule
& \multicolumn{2}{c}{$\mathrm{Fwd}_{\mathrm{by}}$} & \multicolumn{2}{c}{$\mathrm{Fwd}_{\mathrm{major}}$} & \textsc{MechSel} & \multicolumn{2}{c}{\textsc{NEPP}} & \multicolumn{2}{c}{\textsc{Condition}} & \multicolumn{2}{c}{\textsc{Retro}} \\
\cmidrule(lr){2-3}\cmidrule(lr){4-5}\cmidrule(lr){6-6}\cmidrule(lr){7-8}\cmidrule(lr){9-10}\cmidrule(lr){11-12}
\textbf{Method} &
\twolineheight FTS$\uparrow$ & \twolineheight Top-1 &
\twolineheight FTS$\uparrow$ & \twolineheight Top-1 &
\twolineheight Acc. &
\twolineheight FTS$\uparrow$ & \twolineheight Top-1 &
\twolineheight FTS$\uparrow$ & \twolineheight Top-1 &
\twolineheight FTS$\uparrow$ & \twolineheight Top-1 \\
\midrule
ChemOPD, Specialist-only & 0.033 & 0.00 & 0.398 & 0.17 & \textbf{0.45} & 0.462 & \textbf{0.13} & \textbf{0.028} & \textbf{0.03} & 0.437 & 0.01 \\
ChemOPD, no warm-up & 0.034 & 0.00 & 0.396 & 0.15 & 0.40 & 0.505 & 0.11 & 0.000 & 0.00 & 0.303 & 0.00 \\
\rowcolor{chemopdblue!20}\textbf{ChemOPD} & \textbf{0.051} & 0.00 & \textbf{0.465} & \textbf{0.18} & 0.39 & \textbf{0.521} & \textbf{0.13} & 0.023 & \textbf{0.03} & \textbf{0.443} & \textbf{0.03} \\
\bottomrule
\end{tabular}

\endgroup
\end{table*}

\FloatBarrier

\section{Headroom Recovery and Capability Development}
\label{app:capability-analysis}

\subsection{Headroom Recovery}
\label{app:headroom}

Headroom recovery relates the student improvements in
Section~\ref{sec:analysis-opd} to the advantage offered by each routed
specialist. Let $j$ index the reported metric columns and $c_j$ the associated
subtask. Define the direction-aligned utility $U_j(\pi)=s_jm_j(\pi)$, where
$m_j(\pi)$ is the reported score, $s_j=-1$ for MAE, and $s_j=1$ otherwise.
For student method $a$, routed-teacher headroom $H_j$, student gain
$G_j^{(a)}$, and normalized recovery $\mathrm{HR}_j^{(a)}$ are
\begin{equation}
H_j=U_j(T_{r(c_j)})-U_j(\pi_{\theta_0}),\qquad
G_j^{(a)}=U_j(\pi_{\theta}^{(a)})-U_j(\pi_{\theta_0}),\qquad
\mathrm{HR}_j^{(a)}=\frac{G_j^{(a)}}{H_j}.
\label{eq:headroom}
\end{equation}
A negative recovery indicates regression from the common all-task SFT
initializer. A value of one matches the routed teacher's observed gain,
and a value above one exceeds it. Figure~\ref{fig:opd-headroom-recovery}(a--b)
retains every reported metric column. The ratio summary uses the 26 columns
with $H_j>0.02$ to avoid near-zero denominators. 

Positive recovery occurs on 25 of these columns for ChemOPD and 20 for MOPD.
The reported medians are 0.62 and 0.37, respectively. Family medians for MOPD
and ChemOPD are 0.11 and 0.51 for editing/understanding, 0.28 and 0.64 for
optimization, and 0.53 and 0.68 for reaction. These comparisons keep the
specialist checkpoints and routes fixed, so a stronger teacher collection
cannot explain the difference. Recovery is nonetheless a descriptive
normalization of student improvement, not a causal measure of how much
specialist-specific knowledge was transferred. Metric columns are not
independent repeated trials.

  \subsection{Capability Trajectories and Teacher-Relative Signals}
  \label{app:checkpoint-evaluation}

\begin{figure*}[t]
  \centering
  \includegraphics[width=\textwidth]
  {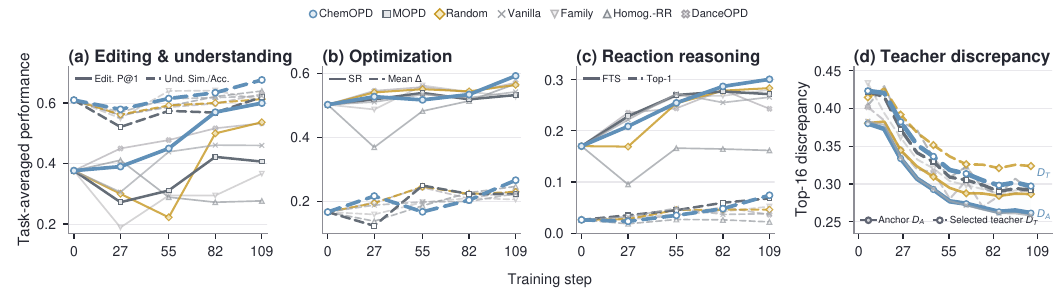}
  \caption{Capability trajectories and teacher-relative signals during OPD
  training. Panels (a--c) report the six task-averaged summaries defined in
  Equation~\ref{eq:capability-summary}. Line style distinguishes the two
  summaries in each panel. Panel (d) reports 10-step means of the anchor and
  selected-specialist discrepancies in
  Equation~\ref{eq:diagnostic-channel-distance}. The discrepancy curves are
  computed on each method's own on-policy prefixes.}
  \label{fig:checkpoint-capability-distance}
  \end{figure*}

  \noindent\textbf{Capability summaries.}
  Figure~\ref{fig:checkpoint-capability-distance} connects capability development
  with the teacher signals observed during OPD. Let $a$ index an OPD method,
  $s\in\{0,27,55,82,109\}$ denote the number of completed actor updates, and
  $\pi_{\theta_s}^{(a)}$ denote the corresponding student. The state at $s=0$ is
  the common all-task SFT initialization. For summary curve $r$, let
  $\mathcal J_r$ contain its task metrics. We report the equal-weight mean
  \begin{equation}
  \overline{M}_{r}^{(a)}(s)
  =
  \frac{1}{|\mathcal J_r|}
  \sum_{j\in\mathcal J_r}
  m_j\!\left(\pi_{\theta_s}^{(a)}\right),
  \label{eq:capability-summary}
  \end{equation}
  where $m_j(\pi)$ is the reported value of metric $j$ for model $\pi$.
  Panels~(a--c) show six summaries: editing P@1 over three tasks;
  understanding similarity or accuracy over three tasks; optimization success
  rate and property change over six properties; and reaction FTS and Top-1 over
  five molecular-output tasks. The understanding summary excludes the two
  count-error metrics, while the reaction summary excludes mechanism selection.
  Because the optimization property changes use different native units, their
  mean is used only to summarize the training trajectory.

  \noindent\textbf{Teacher-relative signals.}
  For update $s\in\{1,\ldots,109\}$, let $i$ index rollout sequences, $\ell$
  index response positions, and $m_{si\ell}\in\{0,1\}$ denote the valid-response
  mask. Let $p^S_{si\ell}$ and $p^{T_{c,i}}_{si\ell}$ be the full-vocabulary
  student and teacher distributions at temperature $\tau=0.7$. The channel
  $c\in\{A,T\}$ denotes the anchor or selected specialist, with
  $T_{A,i}=T_A$ and $T_{T,i}=T_{g(i)}$. Let $v_{si\ell k}$ be the token at rank
  $k$ in the student top-$K$ support. Its normalized student weight is
  \begin{equation}
  q^S_{si\ell k}
  =
  \frac{p^S_{si\ell}(v_{si\ell k})}
  {\sum_{j=1}^{K}p^S_{si\ell}(v_{si\ell j})}.
  \label{eq:diagnostic-topk-weight}
  \end{equation}
  With $N_s=\sum_{i,\ell}m_{si\ell}$, the channel-wise discrepancy is
  \begin{equation}
  D_{s,c}^{(K)}
  =
  \frac{1}{N_s}
  \sum_{i,\ell}m_{si\ell}
  \sum_{k=1}^{K}q^S_{si\ell k}
  \left[
  \log p^S_{si\ell}(v_{si\ell k})
  -
  \log p^{T_{c,i}}_{si\ell}(v_{si\ell k})
  \right],
  \label{eq:diagnostic-channel-distance}
  \end{equation}
  where $K=16$. Panel~(d) reports the mean over each non-overlapping 10-step bin:
  \begin{equation}
  \mathcal B_b
  =
  \{10(b-1)+1,\ldots,\min(10b,109)\},
  \qquad
  \overline{D}_{b,c}^{(16)}
  =
  \frac{1}{|\mathcal B_b|}
  \sum_{s\in\mathcal B_b}D_{s,c}^{(16)}.
  \label{eq:diagnostic-channel-bin}
  \end{equation}
  To interpret this quantity, consider one response position and define
  \begin{equation}
  Z^S=\sum_{k=1}^{K}p^S(v_k),\qquad
  Z^{T_c}=\sum_{k=1}^{K}p^{T_c}(v_k),\qquad
  q^{T_c}_k=\frac{p^{T_c}(v_k)}{Z^{T_c}}.
  \end{equation}
  The token-level discrepancy satisfies
  \begin{equation}
  \sum_{k=1}^{K}q^S_k
  \log\frac{p^S(v_k)}{p^{T_c}(v_k)}
  =
  D_{\mathrm{KL}}\!\left(q^S\,\|\,q^{T_c}\right)
  +
  \log\frac{Z^S}{Z^{T_c}}.
  \label{eq:diagnostic-channel-kl-decomposition}
  \end{equation}
  The support-mass term can be positive or negative. Therefore,
  $D_{s,c}^{(16)}$ is a signed top-$K$ discrepancy rather than a metric
  distance.

  ChemOPD improves all six summaries from the common initialization and achieves
  the highest final value on every curve. The comparison with MOPD is especially
  informative because both methods use the same initialization, specialists,
  routes, prompt stream, and update budget. At update 109, ChemOPD improves the
  editing and understanding summaries from $0.407$ and $0.620$ to $0.600$ and
  $0.677$ relative to MOPD. Optimization SR/$\Delta$ increase from
  $0.532/0.222$ to $0.592/0.266$, while reaction FTS/Top-1 increase from
  $0.272/0.068$ to $0.301/0.074$. MOPD improves the reaction summaries more
  quickly early in training, whereas ChemOPD maintains a higher understanding
  summary and finishes above MOPD on both reaction summaries. This progression
  matches the intended schedule of retaining shared supervision while gradually
  introducing specialist guidance. The ChemOPD anchor discrepancy decreases from
  $0.380$ to $0.262$, and its specialist discrepancy decreases from $0.423$ to
  $0.297$. Across all bins, these curves differ from their Vanilla OPD and MOPD
  counterparts by at most $0.006$ and $0.011$, respectively. ChemOPD therefore
  retains both teacher-relative signal profiles while producing a stronger and
  more balanced capability profile. These results support scheduled
  anchor-residual supervision as an effective way to integrate heterogeneous
  specialist guidance into one student.

\FloatBarrier

 \section{OPD Training Dynamics and Diagnostics}
  \label{app:opd-training-diagnostics}

  This section examines how scheduled anchor-residual supervision changes the
  teacher-derived training signal and the resulting actor updates. The main
  comparison is ChemOPD versus MOPD. They use the same student initialization,
  specialists, routes, prompt stream, optimizer, and update budget. MOPD uses the
  routed specialist directly, whereas ChemOPD combines it with the all-task
  anchor using the schedule in Equation~\ref{eq:alpha-schedule}. This comparison
  therefore tests how the supervision rule affects specialist integration. The
  remaining methods provide additional references for the scale of the training
  signals and updates.

  \noindent\textbf{Training signal and actor update.}
  We use the student top-$K$ support and normalized weights defined in
  Equation~\ref{eq:diagnostic-topk-weight}, with $K=16$. Let $T_i$ denote the
  teacher applied to rollout sequence $i$, and let $\operatorname{sg}$ denote
  stop-gradient. The detached candidate-level advantage is
  \begin{equation}
  \hat A^{(T_i)}_{si\ell k}
  =
  -\operatorname{sg}\!\left(q^S_{si\ell k}\right)
  \operatorname{sg}\!\left[
  \log p^S_{si\ell}(v_{si\ell k})
  -
  \log p^{T_i}_{si\ell}(v_{si\ell k})
  \right].
  \label{eq:diagnostic-topk-advantage}
  \end{equation}
  MOPD sets $T_i=T_{g(i)}$. ChemOPD computes the anchor and routed-specialist
  advantages separately and combines them according to
  Equation~\ref{eq:anchor-residual}. The anchor supplies shared supervision,
  while the scheduled residual adds task-specific corrections.

  Let $\ell^\theta_{si\ell k}$ be the current-student log-probability of
  candidate $v_{si\ell k}$ and let
  $\ell^{\mathrm{old}}_{si\ell k}
  =\operatorname{sg}(\ell^\theta_{si\ell k})$. The actor probability ratio is
  \begin{equation}
  r_{si\ell k}
  =
  \exp\!\left[
  \operatorname{clip}\!\left(
  \ell^\theta_{si\ell k}-\ell^{\mathrm{old}}_{si\ell k},
  -20,20
  \right)
  \right].
  \end{equation}
  With PPO threshold $\epsilon=0.2$ and dual-clip constant $\kappa=3$, the loss
  term for each candidate is
  \begin{align}
  u(\hat A,r)
  &=
  \max\!\left\{
  -\hat A r,\,
  -\hat A\operatorname{clip}(r,1-\epsilon,1+\epsilon)
  \right\},\\
  \psi(\hat A,r)
  &=
  \begin{cases}
  \min\{-\kappa\hat A,u(\hat A,r)\}, & \hat A<0,\\
  u(\hat A,r), & \hat A\geq0.
  \end{cases}
  \end{align}
  The actor loss sums $\psi$ over the $K$ candidates and takes a masked mean over
  valid response positions. Losses from packed microbatches are weighted by
  their numbers of sequences before gradient accumulation. Each reported run
  uses one actor epoch and one logical minibatch per update. Therefore, $r=1$
  when an update is formed, and the local gradient is
  $-\hat A\nabla_\theta\log p_\theta$. No external correctness reward or
  additional KL penalty is used.

  \noindent\textbf{Training diagnostics.}
  We use the update, response-mask, and 10-step-bin notation introduced in
  Appendix~\ref{app:checkpoint-evaluation}. Let $f_i$ denote the benchmark family
  of sequence $i$. For model $q\in\{S,T\}$, its response-token-weighted entropy
  for family $f$ at update $s$ is
  \begin{equation}
  \overline{\mathcal H}^{\,q}_{s,f}
  =
  \frac{
  \sum_{i:f_i=f}\sum_{\ell}m_{si\ell}
  \mathcal H(p^q_{si\ell})
  }{
  \sum_{i:f_i=f}\sum_{\ell}m_{si\ell}
  },
  \qquad
  \mathcal H(p)=-\sum_{v\in\mathcal V}p(v)\log p(v).
  \label{eq:diagnostic-entropy}
  \end{equation}
  Panel~(a) of Figure~\ref{fig:opd-training-dynamics} reports
  \begin{equation}
  E_b
  =
  \frac{1}{|\mathcal F_b|}
  \sum_{f\in\mathcal F_b}
  \frac{1}{|\mathcal B_{b,f}|}
  \sum_{s\in\mathcal B_{b,f}}
  \left|
  \overline{\mathcal H}^{\,S}_{s,f}
  -
  \overline{\mathcal H}^{\,T}_{s,f}
  \right|,
  \label{eq:diagnostic-entropy-gap}
  \end{equation}
  where $\mathcal F_b$ contains the families observed in bin $b$, and
  $\mathcal B_{b,f}\subseteq\mathcal B_b$ contains their updates. This
  macro-average gives each observed family equal weight. For ChemOPD, $T$ is the
  routed MIP specialist rather than the anchor-specialist combination.

  Panel~(b) reports the signed aggregate top-$K$ signal
  \begin{equation}
  S_s^{(K)}
  =
  -\frac{1}{N_s}
  \sum_{i,\ell}m_{si\ell}
  \sum_{k=1}^{K}\hat A_{si\ell k}.
  \label{eq:diagnostic-effective-signal}
  \end{equation}
  For ChemOPD, $\hat A$ is the combined anchor-residual advantage. The negative
  sign follows the plotting convention. This quantity measures the signal
  passed to the actor; it is neither a gradient norm nor a full-vocabulary KL
  divergence. Panel~(c) reports the percentage of valid
  response-position/candidate pairs with large advantages:
  \begin{equation}
  R_s^{\mathrm{ext}}
  =
  100\,
  \frac{
  \sum_{i,\ell}m_{si\ell}
  \sum_{k=1}^{K}
  \mathbf{1}\!\left[|\hat A_{si\ell k}|>5\right]
  }{
  K N_s
  }.
  \label{eq:diagnostic-extreme-rate}
  \end{equation}
  Panels~(a--c) average these quantities within each $\mathcal B_b$. The
  relative reductions annotated in the figure are computed from all 109
  update-level values before binning.

  Let $G_s$ be the total actor-gradient $\ell_2$ norm measured before clipping.
  Panel~(d) shows its empirical CDF and its post-warm-up exceedance rate:
  \begin{equation}
  \widehat F_G(x)
  =
  \frac{1}{109}
  \sum_{s=1}^{109}\mathbf{1}[G_s\leq x],
  \qquad
  C_{\mathrm{post}}
  =
  \frac{1}{103}
  \sum_{s=7}^{109}\mathbf{1}[G_s>1].
  \label{eq:diagnostic-grad-summary}
  \end{equation}
  The second statistic excludes the first six learning-rate warm-up updates.
  The value $1$ is the shared gradient-clipping threshold.

  ChemOPD has a mean student--specialist entropy gap of $0.0280$, compared with
  $0.0313$ for MOPD. This is a reduction of $10.6\%$ and shows closer agreement
  in predictive sharpness. The aggregate top-$K$ signal shows the effect of the
  schedule more directly. In the first 10-step bin, the ChemOPD signal is
  $0.381$, close to anchor-only Vanilla OPD at $0.383$ and below
  specialist-only MOPD at $0.422$. As $\alpha_g(s)$ increases from $0$ to $0.5$,
  the ChemOPD curve moves toward the MOPD curve but remains lower in all 11 bins.
  Across all updates, the mean signal is $0.313$ for ChemOPD and $0.334$ for
  MOPD, a reduction of $6.2\%$.

  The extreme-candidate rate decreases from $0.0314\%$ for MOPD to $0.0222\%$
  for ChemOPD. This $29.4\%$ reduction is observed in every bin. After
  learning-rate warm-up, $36.9\%$ of ChemOPD updates exceed the gradient-clipping
  threshold, compared with $47.6\%$ for MOPD and $35.9\%$ for Vanilla OPD.
  ChemOPD therefore begins with an anchor-like update profile and introduces
  specialist corrections without reaching the larger signal tails of
  specialist-only training. Together with the capability gains in
  Section~\ref{sec:analysis-opd}, these results show that the stronger ChemOPD
  student is not obtained by simply increasing the correction scale. Instead,
  the scheduled anchor-residual objective integrates specialist guidance
  through more moderate updates while retaining broad chemical capability.
\FloatBarrier

\section{Chemistry-Level Case Studies}
\label{app:case-studies}

\begin{figure*}[t]
\centering
\includegraphics[width=\textwidth]{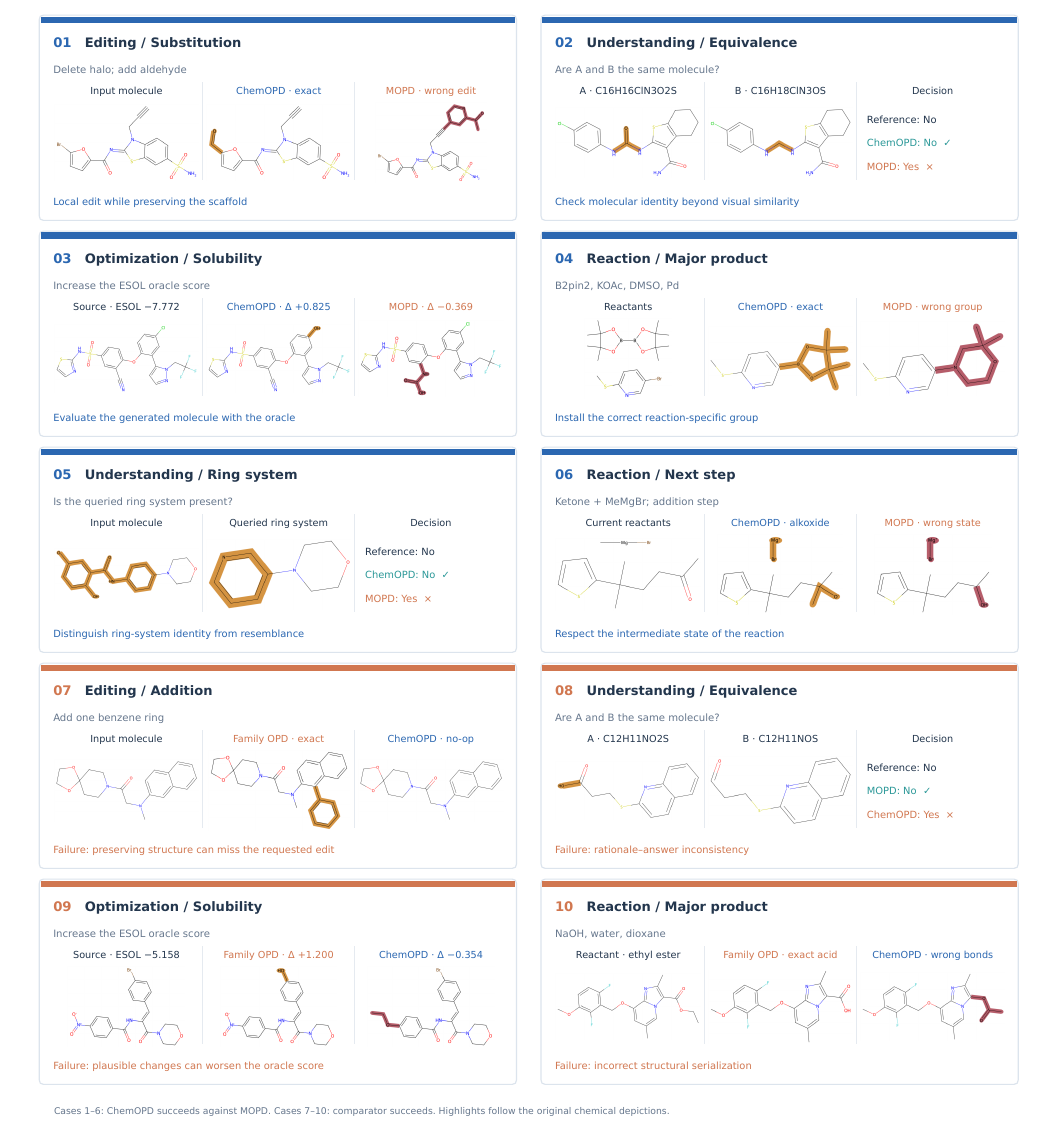}

  \caption{Representative test-set examples across the four chemical task
  families. Cases~1--6, marked with blue headers, show examples in which ChemOPD
  produces the better evaluated result than MOPD. Cases~7--10, marked with
  orange headers, show examples in which the comparison method performs better;
  Case~8 uses MOPD, while the other three use benchmark-family OPD. Orange
  highlights indicate reference-consistent structural features, and rose
  highlights indicate incorrect structural changes. The reported ESOL scores
  are calculated by the same property evaluator used for benchmark scoring.}
\label{fig:chemistry-cases-all}
\end{figure*}

Figure~\ref{fig:chemistry-cases-all} presents ten representative examples
  drawn from the held-out test sets. They cover a broad range of tasks across
  all four task families and illustrate the quality of the capabilities recovered
  by the unified student. The cases were selected for task coverage and chemical
  interpretability. Among
  these examples, ChemOPD produces the stronger task-level result in six cases,
  while the comparison method performs better in the remaining four. We examine whether the final outputs satisfy the task-specific chemical
constraints and use representative responses to illustrate the types of
capabilities evaluated by the benchmark.
\clearpage
\textbf{Complete prompts and generated responses.}
For each case below, the input is shared because the two checkpoints were
evaluated with the same prompt. The response blocks reproduce every
generated output and the final answer.

\medskip
\noindent\textbf{Case 1: Local molecular substitution.}
ChemOPD returns the exact reference molecule and passes the substitution
check. MOPD returns a valid molecule but retains the halo group and fails the
requested edit.

\noindent\textbf{Complete stored evaluation prompt (shared).}
\begin{lstlisting}[style=caseprompt]
"You are a chemical assistent. Given the SMILES structural formula of a molecule, help me ADD and DELETE specified functional groups and output the modified SMILES sequence of the molecule. Input: Molecule SMILES string, Functional Group Names. Output: Modified Molecule SMILES string.
Your response must be directly parsable JSON format:
{
    "output": "Modified Molecule SMILES"
}
Input Molecule: C#CCn1/c(=N/C(=O)c2ccc(Br)o2)sc2cc(S(N)(=O)=O)ccc21, Functional Group to delete: halo, Functional Group to add: aldehyde
\end{lstlisting}

\noindent\textbf{Complete ChemOPD generated response.}
\begin{lstlisting}[style=casechemopd]
{
  "input_structure": "C#CCn1/c(=N/C(=O)c2ccc(Br)o2)sc2cc(S(N)(=O)=O)ccc21",
  "delete_analysis": "The input molecule contains a bromine atom (halo group) attached to a benzene ring in the substituent 'c2ccc(Br)o2'.",
  "add_analysis": "An aldehyde group (-CHO) can be added to the benzene ring at the position previously occupied by the bromine atom, converting it to a benzaldehyde (C=O).",
  "modification": "Replace the bromine atom (Br) with a formyl group (-CHO) on the benzene ring.",
  "output": "C#CCn1/c(=N/C(=O)c2ccc(C=O)o2)sc2cc(S(N)(=O)=O)ccc21"
}
\end{lstlisting}

\noindent\textbf{Complete MOPD generated response.}
\begin{lstlisting}[style=casebaseline]
{
  "input_structure": "C#CCn1/c(=N/C(=O)c2ccc(Br)o2)sc2cc(S(N)(=O)=O)ccc21",
  "delete_strategy": "Remove the bromine atom (halo group) from the aromatic ring at position 4 of the benzene ring.",
  "add_strategy": "Introduce an aldehyde group (-CHO) at the terminal position of the propyl chain attached to the nitrogen atom.",
  "output": "CC(=O)c1cc(C#CCn2/c(=N/C(=O)c3ccc(Br)o3)sc3cc(S(N)(=O)=O)ccc32)ccn1"
}
\end{lstlisting}

\medskip
\noindent\textbf{Case 2: Molecular equivalence.}
The reference label is ``No.'' ChemOPD returns ``No'' and is scored correct;
MOPD returns ``Yes'' and is scored incorrect.

\noindent\textbf{Complete stored evaluation prompt (shared).}
\begin{lstlisting}[style=caseprompt]
You are a chemical assistent. Given two molecule SMILES, Please Determine whether these two Molecules are the same. Input: Molecule A SMILES string, Molecule B SMILES string. Output: yes / no. 
Your response must be directly parsable JSON format:
{
    "output": "Yes / No"
}
Molecule A: NC(=O)c1c(NC(=O)Nc2ccc(Cl)cc2)sc2c1CCCC2, Molecule B: c12c(sc(NCNc3ccc(cc3)Cl)c2C(=O)N)CCCC1.
\end{lstlisting}

\noindent\textbf{Complete ChemOPD generated response.}
\begin{lstlisting}[style=casechemopd]
{
  "input_structure": "Molecule A: NC(=O)c1c(NC(=O)Nc2ccc(Cl)cc2)sc2c1CCCC2",
  "molecule_A_analysis": "A bicyclic system with a thiazole ring fused to a cyclohexane, containing an amide group (NC=O) attached to a benzene ring with a chlorine substituent and another amide group (NC=O) linked to the benzene.",
  "molecule_B_analysis": "A bicyclic system with a thiazole ring fused to a cyclohexane, containing an amide group (C(=O)N) attached to a benzene ring with a chlorine substituent and another amide group (C(=O)N) linked to the benzene.",
  "output": "No"
}
\end{lstlisting}

\noindent\textbf{Complete MOPD generated response.}
\begin{lstlisting}[style=casebaseline]
{
  "input_structure_A": "NC(=O)c1c(NC(=O)Nc2ccc(Cl)cc2)sc2c1CCCC2",
  "molecule_A_analysis": "Contains a thiazole ring fused to a cyclohexane, with two amide groups (NC=O) attached to the thiazole and a chlorophenyl group (c2ccc(Cl)cc2) via an amide linkage.",
  "input_structure_B": "c12c(sc(NCNc3ccc(cc3)Cl)c2C(=O)N)CCCC1",
  "molecule_B_analysis": "Contains a thiazole ring fused to a cyclohexane, with an amide group (C(=O)N) attached to the thiazole and a chlorophenyl group (c3ccc(cc3)Cl) via an amide linkage. The amide is connected to a secondary amine (NCN) instead of a primary amine.",
  "atom_count_comparison": "Both have 14 C, 3 N, 2 O, and 1 Cl. Atom types and counts match.",
  "functional_group_comparison": "Both have thiazole, amide, and chlorophenyl groups. The connectivity of the amide and chlorophenyl groups is identical, differing only in SMILES permutation (amide position and secondary amine substitution).",
  "output": "Yes"
}
\end{lstlisting}

\medskip
\noindent\textbf{Case 3: Solubility optimization.}
Both outputs are valid and retain the source Murcko scaffold. The ESOL deltas
are $+0.824958$ for ChemOPD and $-0.368879$ for MOPD.

\noindent\textbf{Complete stored evaluation prompt (shared).}
\begin{lstlisting}[style=caseprompt]
You are a chemical assistent, Optimize the Source Molecule to improve the compound's ability to dissolve in water (Solubility) while following a structured intermediate optimization process. 
Your response must be directly parsable JSON format:
{
    "Final Target Molecule": "SMILES",
}

Source Molecule: N#Cc1cc(S(=O)(=O)Nc2nccs2)ccc1Oc1ccc(Cl)cc1-c1ccnn1CC(F)(F)F.
\end{lstlisting}

\noindent\textbf{Complete ChemOPD generated response.}
\begin{lstlisting}[style=casechemopd]
{
  "Structural Analysis of Source Molecule": "The source molecule contains a nitrile group (N#C), a sulfonamide (S(=O)(=O)N), a thiazole ring (c2nccs2), a chlorophenyl group (Oc1ccc(Cl)cc1), and a trifluoromethyl-substituted pyridine (c1ccnn1CC(F)(F)F). The presence of the nitrile and sulfonamide groups may contribute to solubility, but the hydrophobic regions (e.g., chlorophenyl, trifluoromethyl) reduce water solubility.",
  "Property Analysis": "The molecule has moderate solubility due to polar groups (sulfonamide, nitrile), but the hydrophobic regions (chlorophenyl, trifluoromethyl) limit its water solubility. The thiazole ring (c2nccs2) is moderately polar but may not significantly enhance solubility.",
  "Limitation in Source Molecule for Property": "The chlorophenyl group (Oc1ccc(Cl)cc1) is highly hydrophobic, and the trifluoromethyl-substituted pyridine (c1ccnn1CC(F)(F)F) adds to the hydrophobicity. The nitrile group (N#C) is polar but may not compensate enough for the hydrophobic regions.",
  "Optimization for Source Molecule": "Replace the chlorophenyl group (Oc1ccc(Cl)cc1) with a more polar substituent, such as a hydroxyl group (O), to improve water solubility. The trifluoromethyl-substituted pyridine (c1ccnn1CC(F)(F)F) can be retained as it may contribute to activity, but the hydrophobicity of the pyridine ring can be mitigated by introducing additional polar groups elsewhere in the molecule.",
  "Final Target Molecule": "N#Cc1cc(S(=O)(=O)Nc2nccs2)ccc1Oc1ccc(O)cc1-c1ccnn1CC(F)(F)F"
}
\end{lstlisting}

\noindent\textbf{Complete MOPD generated response.}
\begin{lstlisting}[style=casebaseline]
{
  "Structural Analysis of Source Molecule": "The source molecule contains a nitrile group (N#C), a sulfonamide group (S(=O)(=O)N), a thiazole ring (c2nccs2), a chlorophenyl group (Oc1ccc(Cl)cc1), and a trifluoromethyl-substituted pyridine (c1ccnn1CC(F)(F)F). The presence of the nitrile and sulfonamide groups may contribute to solubility, but the aromatic systems and halogen substituents reduce water solubility.",
  "Property Analysis": "The molecule has moderate solubility due to the sulfonamide and nitrile groups, but the aromatic rings and halogens (Cl, F) limit its water solubility. The trifluoromethyl group (CC(F)(F)F) is hydrophobic and further reduces solubility.",
  "Limitation in Source Molecule for Property": "The main limitations are the hydrophobic aromatic systems (chlorophenyl, pyridine with trifluoromethyl), the nitrile group (which is polar but may not fully compensate for hydrophobicity), and the lack of additional polar or ionizable groups.",
  "Optimization for Source Molecule": "To improve solubility, we can: 1) Replace the nitrile with a more polar group (e.g., hydroxyl or amine), 2) Introduce a polar substituent (e.g., hydroxyl or amine) on the aromatic rings, 3) Replace the trifluoromethyl group with a more polar group (e.g., hydroxyl or amine), and 4) Add a hydrophilic group (e.g., carboxylic acid or amine) to the pyridine ring.",
  "Final Target Molecule": "O=C(O)Nc1cc(S(=O)(=O)Nc2nccs2)ccc1Oc1ccc(Cl)cc1-c1ccnn1CC(F)(F)F"
}
\end{lstlisting}

\medskip
\noindent\textbf{Case 4: Miyaura borylation product.}
ChemOPD's final molecule exactly matches the reference, whereas MOPD's valid
molecule does not. ChemOPD's reaction-class and solvent descriptions remain
incorrect, as retained below.

\noindent\textbf{Complete stored evaluation prompt (shared).}
\begin{lstlisting}[style=caseprompt]
Given the reactants SMILES and reagents SMILES, your task is to predict the main product SMILES using your experienced chemical Reaction Prediction knowledge.

Input:
Reactants SMILES separated by '.'
Reagents SMILES separated by '.'

Output:
Main product SMILES
[Optional: Byproduct SMILES separated by '.']

The output SMILES must be valid and chemically reasonable. Do not provide any additional information beyond the requested SMILES strings. The answer should be a json format that includes the main product SMILES and byproduct SMILES if any:

Question:
Reactants: CC1(C)OB(B2OC(C)(C)C(C)(C)O2)OC1(C)C.CSc1ccc(Br)cn1
Reagents: CC(=O)O[K].CS(C)=O.[Pd]

Answer:
{
    "Major Product": ...
    "Byproduct(s)": ...
}
\end{lstlisting}

\noindent\textbf{Complete ChemOPD generated response.}
\begin{lstlisting}[style=casechemopd]
{
  "Parse Reactants/Reagents": "Reactant 1: CC1(C)OB(B2OC(C)(C)C(C)(C)O2)OC1(C)C (Bis(pinacolato)diboron, B2pin2). Reactant 2: CSc1ccc(Br)cn1 (2-bromo-5-(methylthio)pyridine). Reagents: CC(=O)O[K] (Potassium acetate), CS(C)=O (Sulfolane, a solvent).",
  "Identify Functional Groups": "Reactant 1: Diboron reagent (source of B-B bond). Reactant 2: Aryl bromide (C-Br bond on pyridine ring), methylthio group (-SMe), pyridine ring. Reagents: Potassium acetate (salt), Sulfolane (solvent).",
  "Predict Reaction Type": "Buchwald-Hartwig amination. This is a palladium-catalyzed cross-coupling reaction between an aryl halide (the bromo-pyridine derivative) and a boronic ester (bis(pinacolato)diboron) to form a C-N bond. The reaction is typically performed in the presence of a base (like potassium acetate) and a palladium catalyst (e.g., Pd(PPh3)4).",
  "Construct Product Structure": "The bromine atom on the pyridine ring of CSc1ccc(Br)cn1 is replaced by the 2-(pinacolato)diboron group from B2pin2. The pinacolato group (B(pin)2) is attached to the carbon atom of the pyridine ring that was previously bonded to the bromine. The methylthio group (-SMe) and the pyridine ring structure remain unchanged. The product is 2-(2-(pinacolato)diboron)-5-(methylthio)pyridine.",
  "Final_SMILES": "CSc1ccc(B2OC(C)(C)C(C)(C)O2)cn1"
}
\end{lstlisting}

\noindent\textbf{Complete MOPD generated response.}
\begin{lstlisting}[style=casebaseline]
{
  "Parse Reactants/Reagents": "Reactant 1: CC1(C)OB(B2OC(C)(C)C(C)(C)O2)OC1(C)C (Bis(pinacolato)diboron, B2pin2). Reactant 2: CSc1ccc(Br)cn1 (2-bromo-5-(methylthio)pyridine). Reagents: CC(=O)O[K] (Potassium acetate), CS(C)=O (Sulfonyl chloride), [Pd] (Palladium catalyst).",
  "Identify Functional Groups": "Reactant 1: Diboron reagent (source of B-B bond, B-B bond is the reactive site). Reactant 2: Aryl bromide (C-Br bond on pyridine ring), methylthio group (-SMe), pyridine ring. Reagents: Potassium acetate (salt), Sulfonyl chloride (source of -SO2Cl, which is a good leaving group), Palladium (catalyst).",
  "Predict Reaction Type": "Palladium-catalyzed Buchwald-Hartwig amination. This reaction couples an aryl halide (in this case, an aryl bromide) with an amine (in this case, a pyridine derivative) to form a new carbon-nitrogen bond. The aryl bromide is the electrophile, and the pyridine derivative is the nucleophile. The reaction is typically promoted by a palladium catalyst and a base (potassium acetate is used as a base).",
  "Construct Product Structure": "The reaction involves the formation of a new C-N bond between the pyridine ring of CSc1ccc(Br)cn1 and the carbon atom of the pyridine ring that was previously bonded to the bromine atom. The bromine atom is replaced by the pyridine nitrogen atom. The methylthio group (-SMe) and the pinacolboronate ester group (-Bpin) are expected to remain unchanged under these conditions. The product is 2-(methylthio)-5-(pinacolboronato)pyridine. The SMILES string for the product is CSc1ccc(N2CCOC(C)(C)C2)cn1.",
  "Final_SMILES": "CSc1ccc(N2CCOC(C)(C)C2)cn1"
}
\end{lstlisting}

\medskip
\noindent\textbf{Case 5: Ring-system recognition.}
The reference label is ``No.'' ChemOPD returns ``No'' and is scored correct;
MOPD returns ``Yes'' and is scored incorrect.

\noindent\textbf{Complete stored evaluation prompt (shared).}
\begin{lstlisting}[style=caseprompt]
You are a chemical assistent. Please Determine whether the ring_system_scaffold is in the Molecule. Input: a molecule's SMILES string, a Ring System Scaffold. Output: yes / no.
Definition: The ring system scaffold consists of one or more cyclic (ring-shaped) molecular structures.
Your response must be directly parsable JSON format:
{
    "output": "Yes / No"
}
Input Molecule: O=C(Nc1ccc(N2CCOCC2)cc1)c1cc(Cl)ccc1O, Ring System Structure: c1cncc(N2CCOCC2)c1
\end{lstlisting}

\noindent\textbf{Complete ChemOPD generated response.}
\begin{lstlisting}[style=casechemopd]
{
  "input_structure": "O=C(Nc1ccc(N2CCOCC2)cc1)c1cc(Cl)ccc1O",
  "molecule_structure_analysis": "Contains a benzamide group (amide attached to a benzene ring with a morpholine substituent and a hydroxyl group on a chlorobenzene ring).",
  "scaffold_analysis": "A pyridine ring (c1cncc) with a morpholine substituent (N2CCOCC2).",
  "matching_analysis": "The molecule includes a benzene ring (c1ccc...cc1) with a morpholine substituent, but the scaffold's pyridine ring (c1cncc) is absent. The benzene ring is a non-aromatic cyclic structure, not matching the scaffold's pyridine core.",
  "output": "No"
}
\end{lstlisting}

\noindent\textbf{Complete MOPD generated response.}
\begin{lstlisting}[style=casebaseline]
{
  "input_structure": "O=C(Nc1ccc(N2CCOCC2)cc1)c1cc(Cl)ccc1O",
  "molecule_structure_analysis": "Contains a benzamide group linked to a 4-chlorophenol and a 2,5-dihydro-1,3-dioxolane (morpholine) substituent.",
  "scaffold_analysis": "A pyridine ring (c1cncc) with a morpholine substituent (N2CCOCC2).",
  "matching_analysis": "The molecule includes a pyridine ring (benzamide's aromatic system) connected to morpholine, matching the scaffold's core structure.",
  "output": "Yes"
}
\end{lstlisting}

\medskip
\noindent\textbf{Case 6: Grignard next elementary step.}
ChemOPD emits the reference magnesium-alkoxide state. MOPD emits a protonated
alcohol and omits the carbon introduced by methylmagnesium bromide.

\noindent\textbf{Complete stored evaluation prompt (shared).}
\begin{lstlisting}[style=caseprompt]
We have one typical reaction (reaction class: 'Bromo Grignard + ester reaction', starting reactants are 'CCOC(=O)CCC(C)(C)c1cccs1.C[Mg]Br', reagents are 'CCOCC.[Cl-].C[Mg]Br.[NH4+].O', and reaction condition is 'Reaction').

Here are the previous elementary reaction steps:

Elementary Step 1: 

{
    "reactants": C[Mg]Br.CCOC(=O)CCC(C)(C)c1cccs1,
    "products": CCOC(C)([O-])CCC(C)(C)c1cccs1.[Mg+]Br,
    "step annotation": Addition,
} 

Elementary Step 2: 

{
    "reactants": CCOC(C)([O-])CCC(C)(C)c1cccs1,
    "products": CC[O-].CC(=O)CCC(C)(C)c1cccs1,
    "step annotation": Leaving or workup,
} 



Now, we want to predict the next elementary reaction step. Currently we know the basic information: 
"current_step_info": {
    "reactants": C[Mg]Br.CC(=O)CCC(C)(C)c1cccs1,
    "step annotation": Addition,
}


Under the same reaction condition and reagents, please give me the products of the next step element reaction. Just return the SMILES of prediction. Your response must contains directly parsable JSON format: 

{
    "pred_smi": str
}
\end{lstlisting}

\noindent\textbf{Complete ChemOPD generated response.}
\begin{lstlisting}[style=casechemopd]
Step-1, Identify Reaction Mechanism, The reaction is a Grignard reaction with an ester. The Grignard reagent (C[Mg]Br) acts as a strong nucleophile, attacking the electrophilic carbonyl carbon of the ester. Step-2, Analyze Previous Steps, Step 1 showed the nucleophilic addition of the Grignard reagent to the ester carbonyl, forming a tetrahedral intermediate. Step 2 showed the collapse of this intermediate, resulting in a carboxylate anion and a ketone. The current step involves the same reactants, but the context is the initial addition of the Grignard reagent to the ester. Step-3, Determine Current Reactants Context, The reactants are the Grignard reagent, C[Mg]Br, and the ester, CC(=O)CCC(C)(C)c1cccs1. The Grignard reagent is a strong nucleophile, and the ester is an electrophile. The reaction is an addition of the Grignard reagent to the carbonyl group. Step-4, Apply Mechanistic Logic, The nucleophilic carbon of the Grignard reagent (C[Mg]Br) will attack the electrophilic carbonyl carbon of the ester (CC(=O)CCC(C)(C)c1cccs1). This is a classic nucleophilic addition to a carbonyl group. The pi-bond of the carbonyl will break, and the electrons will move to the oxygen atom, forming a new C-O bond. The magnesium bromide will remain as a counterion. Step-5, Construct Products, The attack of the Grignard reagent on the ester carbonyl carbon results in the formation of a new C-C bond. The original carbonyl group becomes a tertiary alcohol. The product is a magnesium alkoxide intermediate, specifically a magnesium alkoxide salt of the alcohol. The SMILES for this intermediate is CC(C)([O-])CCC(C)(C)c1cccs1.[Mg+]Br. Step-6, Validate Reagents Conditions, The reaction is carried out in a Grignard reaction solvent (like THF or diethyl ether) and with a base (like ammonium chloride or a salt like magnesium bromide). The formation of a magnesium alkoxide intermediate is a standard and expected step in the Grignard addition to an ester. The conditions are consistent with this transformation. Output: <answer> CC(C)([O-])CCC(C)(C)c1cccs1.[Mg+]Br </answer>
\end{lstlisting}

\noindent\textbf{Complete MOPD generated response.}
\begin{lstlisting}[style=casebaseline]
Step-1, Identify Reaction Mechanism, The reaction is a Grignard reaction with an ester. The first step was the nucleophilic addition of the Grignard reagent to the ester carbonyl, forming a tetrahedral intermediate. The current step involves the addition of the Grignard reagent to the newly formed ketone intermediate. Step-2, Analyze Previous Steps, Step 1 showed the addition of the Grignard reagent (C[Mg]Br) to the ester (CCOC(=O)CCC(C)(C)c1cccs1), forming a tetrahedral intermediate (CCOC(C)([O-])CCC(C)(C)c1cccs1) and a magnesium bromide cation ([Mg+]Br). Step 2 showed the collapse of this tetrahedral intermediate, resulting in a carboxylate anion (CC[O-]) and a ketone (CC(=O)CCC(C)(C)c1cccs1). Step-3, Determine Current Reactants Context, The current reactants are the ketone intermediate (CC(=O)CCC(C)(C)c1cccs1) and the Grignard reagent (C[Mg]Br). The ketone is electrophilic at the carbonyl carbon, and the Grignard reagent is a strong nucleophile. Step-4, Apply Mechanistic Logic, The nucleophilic carbanion of the Grignard reagent (C[Mg]Br) will attack the electrophilic carbonyl carbon of the ketone. This is a classic nucleophilic addition to a carbonyl group. The pi-bond of the carbonyl will break, and the electrons will move to the oxygen atom, forming a new C-O bond. Step-5, Construct Products, The attack of the Grignard reagent on the ketone carbonyl carbon results in the formation of a tertiary alcohol. The carbon atom of the original ketone becomes a tertiary alcohol, and the magnesium bromide cation ([Mg+]Br) is released. The product is the tertiary alcohol (CC(O)CCC(C)(C)c1cccs1) and the magnesium bromide cation ([Mg+]Br). Step-6, Validate Reagents Conditions, The reaction is carried out in the presence of a Grignard reagent and a base (NH4+). The formation of a tertiary alcohol is a standard and expected outcome for a Grignard reaction with a ketone, especially if the reaction is not quenched. The magnesium bromide cation is a common byproduct. Output: <answer> CC(O)CCC(C)(C)c1cccs1.[Mg+]Br </answer>
\end{lstlisting}

\medskip
\noindent\textbf{Case 7: Explicit ring addition.}
Benchmark-family OPD returns the exact reference, while ChemOPD copies the
input molecule and fails the requested addition.

\noindent\textbf{Complete stored evaluation prompt (shared).}
\begin{lstlisting}[style=caseprompt]
You are a chemical assistent. Given the SMILES structural formula of a molecule, help me add a specified functional group and output the improved SMILES sequence of the molecule. Input: Molecule SMILES string, Functional Group Name. Output: Modified Molecule SMILES string.
Your response must be directly parsable JSON format:
{
    "output": "Modified Molecule SMILES"
}

Input Molecule: CN(CC(=O)N1CCC2(CC1)OCCO2)c1ccc2ccccc2c1, Functional Group to add: benzene_ring.
\end{lstlisting}

\noindent\textbf{Complete ChemOPD generated response.}
\begin{lstlisting}[style=casechemopd]
{
  "input_structure": "CN(CC(=O)N1CCC2(CC1)OCCO2)c1ccc2ccccc2c1",
  "functional_group": "benzene_ring",
  "output": "CN(CC(=O)N1CCC2(CC1)OCCO2)c1ccc2ccccc2c1"
}
\end{lstlisting}

\noindent\textbf{Complete benchmark-family OPD generated response.}
\begin{lstlisting}[style=casebaseline]
{
  "input_structure": "CN(CC(=O)N1CCC2(CC1)OCCO2)c1ccc2ccccc2c1",
  "functional_group_addition": "The benzene_ring is added as a substituent to the existing benzene ring in the molecule.",
  "add_strategy": "The benzene_ring is appended to the existing benzene ring at the position where the ethyl group (c1ccc2ccccc2c1) is attached, forming a biphenyl system.",
  "feasibility_analysis": "The modification is chemically viable as it involves attaching a benzene ring to an existing aromatic system without disrupting the core structure. The resulting biphenyl system is stable.",
  "output": "CN(CC(=O)N1CCC2(CC1)OCCO2)c1ccc2ccccc2c1c1ccccc1"
}
\end{lstlisting}

\medskip
\noindent\textbf{Case 8: Acid--aldehyde equivalence.}
The reference label is ``No.'' MOPD returns ``No'' and is scored correct;
ChemOPD returns ``Yes'' and is scored incorrect.

\noindent\textbf{Complete stored evaluation prompt (shared).}
\begin{lstlisting}[style=caseprompt]
You are a chemical assistent. Given two molecule SMILES, Please Determine whether these two Molecules are the same. Input: Molecule A SMILES string, Molecule B SMILES string. Output: yes / no. 
Your response must be directly parsable JSON format:
{
    "output": "Yes / No"
}
Molecule A: O=C(O)CCSc1ccc2ccccc2n1, Molecule B: C(=O)CCSc1ccc2ccccc2n1.
\end{lstlisting}

\noindent\textbf{Complete ChemOPD generated response.}
\begin{lstlisting}[style=casechemopd]
{
  "input_structure": "Molecule A: O=C(O)CCSc1ccc2ccccc2n1 (contains a carboxylic acid group, thioether, and a fused bicyclic aromatic system). Molecule B: C(=O)CCSc1ccc2ccccc2n1 (contains a carbonyl group, thioether, and the same fused bicyclic aromatic system).",
  "atom_count_analysis": "Both molecules have 15 carbons, 1 oxygen, 1 sulfur, and 1 nitrogen. The carboxylic acid in A is replaced by a carbonyl in B, but the core structure (thioether + fused aromatic system) remains identical.",
  "functional_group_analysis": "Molecule A has a carboxylic acid (-COOH) group, while Molecule B has a carbonyl (-C=O) group. The functional group differences are due to the substitution of -OH with -C=O, but the connectivity of the thioether and aromatic system is preserved.",
  "output": "Yes"
}
\end{lstlisting}

\noindent\textbf{Complete MOPD generated response.}
\begin{lstlisting}[style=casebaseline]
{
  "input_structure_A": "O=C(O)CCSc1ccc2ccccc2n1",
  "molecule_A_analysis": "A carboxylic acid group (O=C(O)) attached to a propyl chain (CC) connected to a thioether (S) linking to a fused bicyclic aromatic system (benzene fused with a pyridine-like ring).",
  "input_structure_B": "C(=O)CCSc1ccc2ccccc2n1",
  "molecule_B_analysis": "A carboxylic acid group (C(=O)) attached to a propyl chain (CC) connected to a thioether (S) linking to a fused bicyclic aromatic system (benzene fused with a pyridine-like ring).",
  "atom_count_comparison": {
    "A": {
      "C": 10,
      "O": 2,
      "S": 1,
      "N": 1
    },
    "B": {
      "C": 10,
      "O": 1,
      "S": 1,
      "N": 1
    },
    "output": "No",
    "atom_count_mismatch": {
      "O": 1,
      "output": "No",
      "atom_count_difference": {
        "O": 1
      }
    }
  },
  "bond_count_comparison": {
    "A": 10,
    "B": 10,
    "output": "No",
    "bond_count_mismatch": {
      "output": "No",
      "bond_count_difference": {}
    }
  },
  "functional_group_comparison": {
    "A": "Carboxylic acid, thioether, fused bicyclic aromatic system",
    "B": "Carboxylic acid, thioether, fused bicyclic aromatic system",
    "output": "No",
    "functional_group_identicality": "Yes",
    "functional_group_mismatch": "Carboxylic acid group: O=C(O) vs C(=O) in SMILES representation"
  },
  "output": "No"
}
\end{lstlisting}

\medskip
\noindent\textbf{Case 9: Solubility-oracle boundary.}
The ESOL delta is $+1.199869$ for benchmark-family OPD and $-0.354096$ for
ChemOPD; both outputs are valid and retain the source Murcko scaffold.

\noindent\textbf{Complete stored evaluation prompt (shared).}
\begin{lstlisting}[style=caseprompt]
You are a chemical assistent, Optimize the Source Molecule to improve the compound's ability to dissolve in water (Solubility) while following a structured intermediate optimization process. 
Your response must be directly parsable JSON format:
{
    "Final Target Molecule": "SMILES",
}

Source Molecule: O=C(NC(=Cc1ccc(Br)cc1)C(=O)N1CCOCC1)c1ccc([N+](=O)[O-])cc1.
\end{lstlisting}

\noindent\textbf{Complete ChemOPD generated response.}
\begin{lstlisting}[style=casechemopd]
{
  "Structural Analysis of Source Molecule": "The source molecule contains a central amide group (O=C-N) connected to a brominated benzyl group (C(=Cc1ccc(Br)cc1)) and a nitro-substituted benzene ring (c1ccc([N+](=O)[O-])cc1). The nitro group ([N+](=O)[O-]) is highly polar but may limit solubility due to its strong electron-withdrawing nature.",
  "Property Analysis": "The nitro group contributes to high polarity but may reduce water solubility due to its strong electron-withdrawing effect, which can destabilize the molecule. The brominated benzyl group adds hydrophobicity.",
  "Limitation in Source Molecule for Property": "The nitro group ([N+](=O)[O-]) is a major limitation for water solubility due to its high polarity and potential for hydrogen bonding disruption. The bromine atom also adds hydrophobicity.",
  "Optimization for Source Molecule": "Replace the nitro group ([N+](=O)[O-]) with a methoxy group (OCH3) to reduce polarity and improve solubility. The methoxy group is less electron-withdrawing and more hydrophilic, enhancing water solubility while maintaining some aromatic character.",
  "Final Target Molecule": "O=C(NC(=Cc1ccc(Br)cc1)C(=O)N1CCOCC1)c1ccc(OCC)cc1"
}
\end{lstlisting}

\noindent\textbf{Complete benchmark-family OPD generated response.}
\begin{lstlisting}[style=casebaseline]
{
  "Structural Analysis of Source Molecule": "The source molecule contains a central amide group (O=C-N) connected to a brominated benzyl group (C(=Cc1ccc(Br)cc1)) and a nitro-substituted benzene ring (c1ccc([N+](=O)[O-])cc1). The nitro group ([N+](=O)[O-]) is highly polar and electron-withdrawing, while the bromine (Br) is a hydrophobic substituent.",
  "Property Analysis": "The nitro group enhances water solubility due to its polarity and ability to form hydrogen bonds. However, the bromine atom reduces solubility by introducing hydrophobicity. The amide and ether groups (C(=O)N1CCOCC1) contribute to moderate solubility but are insufficient to counteract the hydrophobic effects of Br and the nitro group.",
  "Limitation in Source Molecule for Property": "The bromine atom is the primary limitation for water solubility due to its hydrophobic nature. The nitro group, while polar, is not enough to offset the hydrophobicity introduced by Br.",
  "Optimization for Source Molecule": "Replace the bromine atom with a more polar and hydrophilic group, such as a hydroxyl (-OH) or a carboxyl (-COOH) group, to enhance water solubility. The nitro group can be retained to maintain polarity and hydrogen bonding capacity.",
  "Final Target Molecule": "O=C(NC(=Cc1ccc(O)cc1)C(=O)N1CCOCC1)c1ccc([N+](=O)[O-])cc1"
}
\end{lstlisting}

\medskip
\noindent\textbf{Case 10: Ester saponification.}
Benchmark-family OPD exactly matches the reference carboxylic acid. ChemOPD's
product is valid but has the wrong connectivity and does not match the
reference.

\noindent\textbf{Complete stored evaluation prompt (shared).}
\begin{lstlisting}[style=caseprompt]
Given the reactants SMILES and reagents SMILES, your task is to predict the main product SMILES using your experienced chemical Reaction Prediction knowledge.

Input:
Reactants SMILES separated by '.'
Reagents SMILES separated by '.'

Output:
Main product SMILES
[Optional: Byproduct SMILES separated by '.']

The output SMILES must be valid and chemically reasonable. Do not provide any additional information beyond the requested SMILES strings. The answer should be a json format that includes the main product SMILES and byproduct SMILES if any:

Question:
Reactants: CCOC(=O)c1c(C)nc2c(OCc3c(F)ccc(OC)c3F)cc(C)cn12
Reagents: O.O[Na].C1COCCO1

Answer:
{
    "Major Product": ...
    "Byproduct(s)": ...
}
\end{lstlisting}

\noindent\textbf{Complete ChemOPD generated response.}
\begin{lstlisting}[style=casechemopd]
{
  "Parse Reactants/Reagents": "Reactant 1: CCOC(=O)c1c(C)nc2c(OCc3c(F)ccc(OC)c3F)cc(C)cn12 (an ethyl ester). Reagent 1: O.O[Na].C1COCCO1 (1,4-Dioxane-2,5-diol, a cyclic ether).",
  "Identify Functional Groups": "Reactant 1: Ethyl ester (-COOEt), pyrazolo[1,5-a]pyrimidine core, methyl groups, methoxy groups on a substituted phenyl ring, and a fluorine atom. Reagent 1: 1,4-Dioxane-2,5-diol (a cyclic ether, specifically a 1,4-dioxane derivative, which can act as a nucleophile or a base).",
  "Predict Reaction Type": "The reaction is predicted to be a base-catalyzed ester hydrolysis (saponification). The 1,4-dioxane-2,5-diol (reagent 1) acts as a base, and the ethyl ester (reactant 1) is the substrate. The reaction proceeds via nucleophilic acyl substitution, where the oxygen of the 1,4-dioxane-2,5-diol attacks the carbonyl carbon of the ester, leading to the cleavage of the C-OEt bond and formation of a carboxylate salt and ethanol. Subsequent protonation (typically during aqueous workup) yields the carboxylic acid.",
  "Construct Product Structure": "The ethyl ester group (-C(=O)OCC) is hydrolyzed to a carboxylic acid group (-C(=O)OH). The rest of the molecule, including the pyrazolo[1,5-a]pyrimidine core, the methyl groups, the methoxy groups on the phenyl ring, and the fluorine atom, remains unchanged. The product is the carboxylic acid: CC(=O)Oc1c(C)nc2c(OCc3c(F)ccc(OC)c3F)cc(C)cn12.",
  "Final_SMILES": "CC(=O)Oc1c(C)nc2c(OCc3c(F)ccc(OC)c3F)cc(C)cn12"
}
\end{lstlisting}

\noindent\textbf{Complete benchmark-family OPD generated response.}
\begin{lstlisting}[style=casebaseline]
{
  "Parse Reactants/Reagents": "Reactant 1: CCOC(=O)c1c(C)nc2c(OCc3c(F)ccc(OC)c3F)cc(C)cn12 (an ethyl ester). Reagent 1: O (water). Reagent 2: O[Na] (sodium hydroxide). Reagent 3: C1COCCO1 (1,4-dioxane, a solvent).",
  "Identify Functional Groups": "Reactant 1 contains an ethyl ester group (-C(=O)OCC), a pyrazolo[1,5-a]pyrimidine core, a methoxy group (-OCH3), and a difluoromethoxy group (-OCH2F). The ester group is the most reactive site for the reaction with sodium hydroxide.",
  "Predict Reaction Type": "Saponification (base-catalyzed hydrolysis of an ester). The hydroxide ion (OH-) from sodium hydroxide attacks the electrophilic carbonyl carbon of the ester. This leads to the cleavage of the acyl-oxygen bond, forming a carboxylate salt and ethanol. Subsequent protonation (typically during aqueous workup) yields the carboxylic acid.",
  "Construct Product Structure": "The ethyl ester group (CCOC(=O)-) is converted into a carboxylic acid group (O=C(O)-). The rest of the molecule, including the pyrazolo[1,5-a]pyrimidine core, the methoxy group, and the difluoromethoxy group, remains unchanged. The SMILES fragment CCOC(=O)- becomes O=C(O)-.",
  "Final_SMILES": "O=C(O)c1c(C)nc2c(OCc3c(F)ccc(OC)c3F)cc(C)cn12"
}
\end{lstlisting}

\FloatBarrier

\end{document}